\documentclass{article}

\PassOptionsToPackage{numbers,sort&compress}{natbib}
\usepackage[main,final]{neurips_2026}

\usepackage[utf8]{inputenc} 
\usepackage[T1]{fontenc}    
\usepackage{hyperref}       
\usepackage{url}            
\usepackage{booktabs}       
\usepackage{amsfonts}       
\usepackage{nicefrac}       
\usepackage{microtype}      
\usepackage{graphicx}
\graphicspath{{pictures/}}
\usepackage{amsmath}
\usepackage{multirow}
\usepackage[table]{xcolor}
\usepackage{algorithm}
\usepackage{algorithmic}
\usepackage{wrapfig}

\title{SchemaPose: RGB-Based Category-Level Pose Estimation with Parametric Category Schema}

\author{
  Shuang Wu\thanks{Equal contribution.} \\
  Shanghai Jiao Tong University \\
  \texttt{souglad21@sjtu.edu.cn} \\
  \And
  Jiude Wei\footnotemark[1] \\
  Shanghai Jiao Tong University \\
  \texttt{wjd\_kznwl@sjtu.edu.cn} \\
  \And
  Cewu Lu \\
  Shanghai Jiao Tong University \\
  \texttt{lucewu@sjtu.edu.cn} \\
  \And
  Jianhua Sun\thanks{Corresponding author.} \\
  Shanghai Jiao Tong University \\
  \texttt{gothic@sjtu.edu.cn}
}

\begin{document}

\maketitle

\begin{abstract}
RGB-based category-level 9D pose estimation remains highly challenging because a monocular system must generalize across unseen object instances while jointly inferring 3D translation, 3D rotation, and 3D size. Unlike instance-level pose estimation, the category-level setting requires the network to learn both what is shared across objects within a category and how individual instances vary in structure. This challenge is amplified in the RGB-only setting, where pose cues, category-level commonality, and instance-specific variation are entangled in appearance without direct geometric constraints, making it difficult for existing frameworks to learn a representation that is truly aligned with the category-level estimation objective. To address this problem, we propose SchemaPose, a schema-guided framework for RGB-based category-level pose estimation. SchemaPose equips the estimation pipeline with an explicit parametric schema for modeling category-level regularities together with structured instance variation, and uses the inferred schema variables to guide pose-related prediction in an end-to-end query-based framework. The same formulation also supports template instantiation, synthetic annotation, and pose learning within a unified framework. Experiments on synthetic and real data show that SchemaPose achieves state-of-the-art performance among RGB-based methods for category-level 9D pose estimation.
\end{abstract}

\section{Introduction}

Object pose estimation is a fundamental problem in computer vision, with broad applications in robotic manipulation~\cite{fu20226d,tremblay2018deep,xiang2017posecnn,wang2019densefusion}, augmented reality~\cite{tang20193d,marchand2015pose,kato1999marker}, and embodied perception~\cite{savva2019habitat,xiang2020sapien,kolve2017ai2}. Compared with instance-level pose estimation, category-level pose estimation aims to recover the poses of previously unseen object instances within a predefined category, making it both more practical for real-world deployment and substantially more challenging~\cite{wang2019nocs}. This challenge becomes especially pronounced in the RGB-only setting, where a monocular system must jointly infer 3D rotation, 3D translation, and 3D size from a single image alone, despite large structural variation across object instances within the same category~\cite{huang2025givepose,zhang2024lapose,liu2025monodiff9d}.

A key difficulty of category-level pose estimation is that the network must learn not only pose, but also the category-level structure shared across instances and the ways in which individual objects deviate from it~\cite{huang2025givepose,wan2023socs}. This requirement is especially demanding for RGB-based methods, because without direct geometric constraints from depth, pose cues are tightly entangled with instance-specific shape, part configuration, and object extent in monocular RGB observations~\cite{zhang2024lapose,wei2024dmsr}. Consequently, structural differences across instances can affect the interpretation of 3D size and spatial layout, and further complicate rotation and translation estimation, making it difficult for existing frameworks to learn an internal representation aligned with the category-level pose objective.

A representative line of RGB-based methods~\cite{fan2022oldnet,lee2021msos,wei2024dmsr} builds on NOCS maps~\cite{wang2019nocs}, which provide a shared canonical coordinate space for different instances of the same category. 
Pose is then recovered from the predicted 2D--3D or 3D--3D correspondences using geometric solvers~\cite{umeyama2002least,lepetit2009epnp}. 
However, recent studies~\cite{huang2025givepose,wan2023socs} show that NOCS maps can be suboptimal for category-level direct prediction, since different object instances may produce different NOCS maps under similar category-level poses, making the mapping from canonical coordinates to pose unnecessarily complex. Moreover, the predicted coordinates often retain substantial instance-specific geometric detail that is not directly aligned with stable pose prediction. 
Recent methods~\cite{lee2025yopo,fischer2025unified} therefore move toward more direct end-to-end prediction and avoid explicit canonical maps. However, without an explicit structural variable, pose-relevant category structure and instance variation are typically absorbed implicitly into latent or query features, making them hard to supervise, control, or consistently use across the estimation process. 

This observation suggests that RGB-based category-level pose estimation requires an explicit structural variable that can capture both category-level regularity and instance-level deviation, and can be used directly throughout the estimation process. In cognitive science~\cite{biederman1987recognition,bartlett1995remembering,rumelhart2017schemata}, such structured abstractions are often described as schemas: internal organizational models that preserve what is stable across a category while accommodating lawful variation across instances. Rather than focusing on the surface details of individual objects, a schema captures the shared structure of a category together with the factors along which instances differ. 

Motivated by this, we propose SchemaPose, a structure-aware framework for RGB-based category-level 9D pose estimation. SchemaPose introduces a parametric category schema as an explicit structural layer in a unified query-based pipeline, allowing category-level regularity and instance-level variation to be modeled and used directly for pose prediction. We develop a dual-stage schema-guided mechanism so that inferred schema participates in both proposal-stage coarse pose prediction and decoder-stage pose refinement, making structural information available throughout the estimation process rather than only as an auxiliary target. Beyond pose prediction, the same parametric formulation supports template instantiation and synthetic data generation, providing structurally consistent supervision for training.

We evaluate SchemaPose on CAMERA25 and REAL275, covering both synthetic and real-world settings. SchemaPose consistently outperforms existing RGB-based category-level pose estimation methods across all metrics. In particular, compared with the strongest prior RGB-based method, SchemaPose improves the $10^\circ$10cm metric from 38.7 to 43.1 on CAMERA25 and from 52.8 to 68.4 on REAL275. Further ablation studies verify that the strong performance of SchemaPose benefits from using meaningful schema information as an explicit structural condition for pose decoding, rather than relying solely on implicit visual features to absorb category-level structural variation.

To summarize, our contributions are as follows.
\textbf{First,} 
we propose SchemaPose, a schema-guided framework for RGB-based category-level 9D pose estimation, enabling the estimation pipeline to better learn category-level commonality and instance variation for pose prediction. 
\textbf{Second,}
we introduce a parametric category schema to explicitly model category-level structure and instance variation, and build a schema-driven data generation pipeline to provide accurate synthetic supervision for SchemaPose training.
\textbf{Third,} 
we develop a dual-stage schema-guided mechanism, where schema inference participates from proposal generation to decoder-level pose prediction, allowing category-level structural information to directly influence the full pose estimation process.
\textbf{Finally,} we validate the proposed framework on synthetic and real benchmarks, where SchemaPose surpasses the strongest prior RGB-based method by a large margin, especially on real-world pose accuracy.

\section{Related Work}

\subsection{RGB-Based Category-Level Pose Estimation}

RGB-based category-level pose estimation aims to recover poses of unseen object instances from monocular images without requiring instance-specific CAD models at inference time. A representative line of work remains geometry-guided. NOCS~\cite{wang2019nocs} establishes a shared canonical space and recovers pose through canonical correspondences, forming the basis of many later RGB-based extensions. MSOS~\cite{lee2021msos} jointly predicts metric-scale shape and NOCS and renders depth for similarity alignment. OLD-Net~\cite{fan2022oldnet} estimates object-level depth together with canonical coordinates and recovers pose through geometric alignment. DMSR~\cite{wei2024dmsr} further decouples metric-scale recovery from rigid pose recovery and solves pose from learned correspondences via PnP.

Recent methods increasingly move toward direct regression or more unified prediction. LaPose~\cite{zhang2024lapose} performs end-to-end pose regression with uncertainty-aware NOCS modeling, while GIVEPose~\cite{huang2025givepose} argues that instance-specific NOCS maps are suboptimal for direct category-level regression and progressively transforms NOCS toward IVFC. YOPO~\cite{lee2025yopo} formulates category-level 9D pose estimation as DETR-style joint detection and pose prediction from object queries without an explicit alignment stage. LanCOPE~\cite{yang2025lancope} further introduces language-guided multimodal fusion for monocular pose prediction, and Fischer et al.~\cite{fischer2025unified} propose a unified RGB-only framework based on 3D prototypes, dense 2D/3D matching, and multi-model RANSAC PnP. Despite their differences, existing methods either organize pose estimation around canonical geometry and explicit geometric solving, or absorb pose-relevant structural factors implicitly into learned features. Our method instead introduces schema as an explicit structural variable inside the pose estimation framework.

\subsection{Canonical Coordinates and Intermediate Representations}

Canonical coordinates are a central design choice in category-level pose estimation because they provide a shared reference space for unseen instances. NOCS~\cite{wang2019nocs} is the most influential formulation and serves as the basis for many subsequent methods. Recent studies refine this line from different perspectives. SOCS~\cite{wan2023socs} improves semantic consistency under large intra-category variation through keypoint-guided non-rigid warping. LaPose~\cite{zhang2024lapose} models uncertainty over canonical coordinates under monocular ambiguity. GIVEPose~\cite{huang2025givepose} progressively transforms NOCS toward IVFC for direct regression, and DiffusionNOCS~\cite{ikeda2024diffusionnocs} keeps NOCS as the target while replacing deterministic prediction with diffusion-based estimation under multi-modal inputs.

These methods improve pose estimation mainly by redesigning the canonical space, the uncertainty model, or the prediction process of canonical maps. In contrast, our method does not pursue another coordinate-space variant. Instead, it introduces schema-guided structural abstraction as a framework-level conditioning variable, aiming to explicitly represent category-level commonality and instance-level variation in a form that can directly support pose prediction.

\section{Method}
\label{headings}

\subsection{Overview}

Given an RGB image \(I\) and the camera intrinsic matrix \(K \in \mathbb{R}^{3\times 3}\), our goal is to estimate the category-level 9D pose of all object instances in the image. For the set of object instances \(\{o_i\}_{i=1}^{N}\), the network predicts
\[
\{(c_i, s_i, R_i, t_i)\}_{i=1}^{N}
\]
where \(c_i \in \mathcal{C}\) is the category label from the predefined category set \(\mathcal{C}\), \(s_i \in \mathbb{R}^{3}\) denotes the 3D size, \(R_i \in SO(3)\) denotes the 3D rotation, and \(t_i \in \mathbb{R}^{3}\) denotes the 3D translation of the \(i\)-th object instance.

Unlike common two-stage pipelines~\cite{lee2021msos,wei2024dmsr,liu2025monodiff9d,zhang2024lapose} that first apply an off-the-shelf detector
(e.g., Mask R-CNN~\cite{he2017maskrcnn}) to localize object regions and then
estimate pose from cropped object images, \textbf{SchemaPose} is built on the
end-to-end object detector DINO~\cite{zhang2022dino} and extends it to
category-level pose estimation with explicit structural guidance. Specifically,
SchemaPose integrates object detection, schema inference, and pose-related
prediction into a unified query-based architecture. Its core component is a
\textbf{dual-stage schema-guided mechanism}, where ``dual-stage'' refers to
schema-guided prediction at the proposal-generation stage and the decoder-refinement
stage within the same end-to-end network, rather than a separate
detection-then-pose pipeline.

To support this framework, we first introduce a \textbf{parametric category
schema} and a \textbf{schema-driven data generation pipeline} in Sec.~\ref{section schema}. The
schema defines the explicit structural variables used for schema-guided pose
prediction, while the generation pipeline provides diverse schema-annotated
object instances and rendered training images with aligned schema and pose
labels. We then present the \textbf{SchemaPose framework} in Sec.~\ref{section framework}, detailing
how the dual-stage schema-guided mechanism enables structural information to
participate throughout the unified detection and pose estimation process.

\begin{figure*}[t]
    \centering
    \includegraphics[trim=0 10 0 0,clip,width=\linewidth]{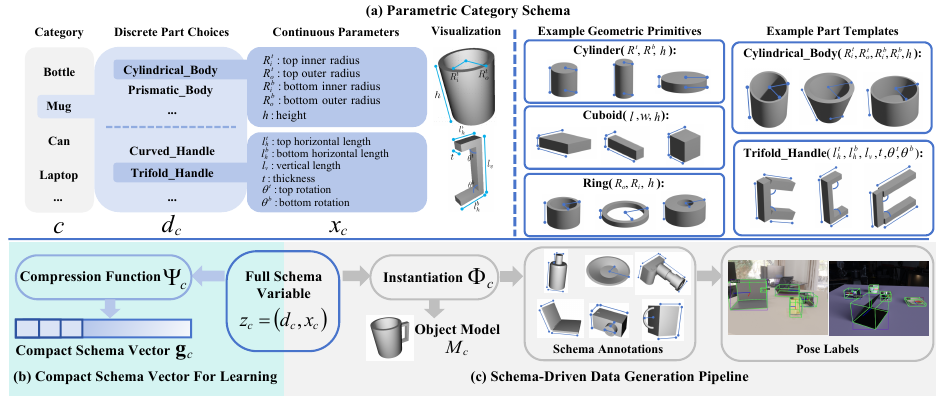}
    \vspace{-12pt}
    \caption{
    {\textbf{Overview of the parametric category schema, compact representation, and data generation pipeline.} \textbf{(a) Parametric Category Schema.} For category $c$, the schema comprises discrete part choices $d_c$ and continuous geometric parameters $x_c$, which define a valid object structure via semantic part templates. \textbf{(b) Compact Representation for Learning.} A category-specific compression function $\Psi_c$ maps the full schema variable $z_c=(d_c,x_c)$ to a compact vector $\mathbf{g}_c$, preserving the dominant pose-relevant structural factors for network training. \textbf{(c) Schema-Driven Data Generation.} The schema variable $z_c$ is analytically instantiated into an object model $M_c$ via $\Phi_c$. These models are diversified through parameter perturbation and rendered within synthetic scenes to generate training images with aligned ground-truth pose labels.}
    }
    \vspace{-9pt}
    \label{fig:schema}
\end{figure*}

\begin{wrapfigure}{r}{0.5\linewidth} 
    \centering
    \vspace{-15pt} 
    \includegraphics[trim=0 30 50 0,clip,width=\linewidth]{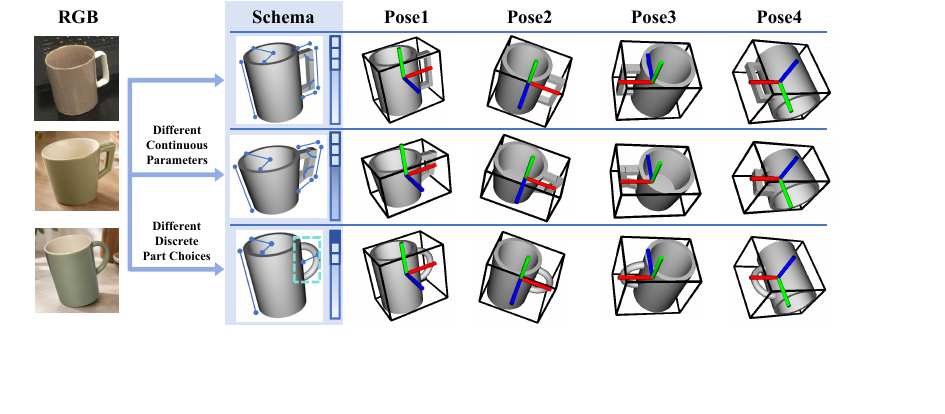}
    \vspace{-18pt} 
    \caption{
    {Relationship between object schema and pose. Objects are parameterized through schema variables, which provide the structural basis for expressing and interpreting 9D pose. }
    }
    \label{fig:schema_pose}
    \vspace{-12pt} 
\end{wrapfigure}

\subsection{Schema Representation and Schema-Driven Data Generation}
\label{section schema}

In SchemaPose, the parametric category schema serves two connected roles: it defines an explicit structural variable for category-level pose learning, and it provides the generative specification for constructing synthetic training data with aligned schema and pose annotations.

\paragraph{Parametric Category Schema.}
We define the schema for category \(c\) as a parameterized abstraction that models object structure using geometric primitives (e.g., cuboids, cylinders, rings) and semantic part templates (e.g., body, handle, lid).
Each object instance is specified by a full schema variable \(z_c=(d_c,x_c)\), i.e., the complete discrete-continuous encoding of its structure under the category schema. The discrete variables \(d_c\) select the active part configuration and template choices, while the continuous variables \(x_c\) specify the geometric attributes of the selected templates, such as radius, thickness, height, offset, or curvature.
Formally, the instance is generated by
\begin{equation}
    M_c = \Phi_c(d_c, x_c),
\end{equation}
where \(\Phi_c\) is the instantiation function that analytically constructs the object by composing the selected part templates and their geometric relations. 
This formulation represents a category as a valid, category-consistent family of object instances rather than a fixed prototype, providing a structural basis for both schema-driven data generation and schema-guided pose learning.
Fig.~\ref{fig:schema}(a) illustrates an example of the schema abstraction.
 
\paragraph{Compact Schema Vector for Learning.}
As shown in Fig.~\ref{fig:schema}(b), we distinguish the full schema variable used for object instantiation from the compact schema vector used for network learning. The full schema variable \(z_c=(d_c,x_c)\) specifies all active part templates and geometric parameters required to analytically construct a valid object instance. For pose estimation, however, directly regressing the full instantiated schema would introduce excessive output dimensionality and template-dependent parameter subsets. We therefore derive, for each category, a compact schema vector
\begin{equation}
    \mathbf{g}_c=\Psi_c(z_c),
\end{equation}
where \(\Psi_c(\cdot)\) is a category-specific compression function.
This function retains pose-relevant structural factors, such as object extent, proportions, and coarse part configuration, while removing low-level geometric details used only for exact instantiation. Continuous parameters are further normalized by category-specific reference scales or valid parameter ranges, and inactive optional-part parameters are zero-padded to form a fixed category-wise layout. As a result, \(\mathbf{g}_c\) provides a compact, normalized, and stable structural target for pose estimation without requiring the network to predict the full instantiable schema. 
More details on schema implementation, including geometric primitive construction, part-template composition, and compact schema compression, are provided in Appendix~\ref{app:schema_implementation}.

\paragraph{Schema-Driven Synthetic Data Generation.}
As shown in Fig.~\ref{fig:schema}(c), our pipeline contains two stages. 
\textbf{First, schema-driven object generation.}
Given a valid full schema variable \(z_c\), we instantiate a category-consistent object by activating the selected part templates and assigning the corresponding continuous parameters. Varying \(z_c\) within category-consistent ranges yields structurally diverse objects. Since each object is generated from known schema parameters, we compute its compact schema vector \(\mathbf{g}_c=\Psi_c(z_c)\) and use it as the schema annotation for network supervision.
\textbf{Second, compositional scene rendering.}
Generated objects are sampled from the object pool, randomly scaled and posed, and placed on supporting surfaces under collision and visibility constraints. We reject placements with severe interpenetration, extreme truncation, or insufficient visible area, and render RGB images in BlenderProc~\cite{bproc} with varying viewpoints, lighting, materials, textures, and backgrounds. Since the full scene state is known, each rendered image provides aligned pose labels and compact schema annotations for training SchemaPose. More implementation details of the data generation pipeline can be found in Appendix~\ref{app:data_generation}.


\begin{figure*}[t]
    \centering
    \includegraphics[trim=0 35 0 0,clip,width=\linewidth]{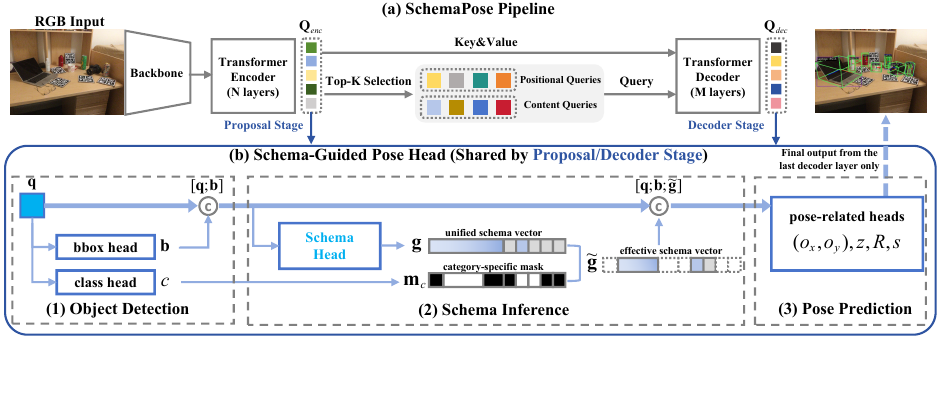}
    \vspace{-12pt}
    \caption{
    Overview of SchemaPose and Dual-Stage Schema Guided Mechanism.
    (a) SchemaPose follows an end-to-end query-based pipeline. Given an RGB image, the backbone and Transformer encoder produce encoder memory \(\mathbf{Q}_{enc}\), from which top-K proposals are selected to initialize decoder queries. The Transformer decoder refines these queries and produces decoder features \(\mathbf{Q}_{dec}\). 
    (b) The Schema-Guided Pose Head (SGPH) follows the same design at the proposal and decoder stages. Given an object feature \(\mathbf{q}\), SGPH first predicts the object category \(c\) and bounding box \(\mathbf{b}\), then infers a unified schema vector \(\mathbf{g}\). A category-specific mask \(\mathbf{m}_c\) selects the effective schema representation \(\tilde{\mathbf{g}}\), which is fed back as structural guidance for pose-related prediction. During training, detection, schema, and pose predictions are supervised at both stages, while inference uses the final pose output \((o_x,o_y), z, R, s\) from the last decoder layer.
    }
    \vspace{-6pt}
    \label{fig:framework}
\end{figure*}

\subsection{SchemaPose Framework}
\label{section framework}

In this section, we present SchemaPose, whose core design comes from the relationship between object schema and pose: as illustrated in Fig.~\ref{fig:schema_pose}, parametric category schema provides a structural basis for category-level pose estimation. Given an RGB image, SchemaPose explicitly infers pose-relevant structural factors in the form of a compact schema vector and uses them to guide pose prediction.

\paragraph{Overall Architecture.}
As shown in Fig.~\ref{fig:framework}, SchemaPose adopts an end-to-end query-based architecture consisting of a backbone, a multi-layer Transformer encoder, a multi-layer Transformer decoder, and a Schema-Guided Pose Head (SGPH). Given an input RGB image, the backbone extracts multi-scale features, which are further enhanced by the Transformer encoder to produce the encoder memory \(\mathbf{Q}_{enc}\). The SGPH is first applied to the encoder memory to produce proposal-level predictions, including object category, bounding box, schema, and pose-related quantities. The top-K proposals are then selected according to their classification scores and used to initialize decoder queries. The Transformer decoder refines these queries through cross-attention with the encoder memory, and the same SGPH is applied to the refined query features \(\mathbf{Q}_{dec}\) at each decoder layer to produce decoder-level predictions.

We refer to this design as a \textbf{Dual-Stage Schema-Guided Mechanism}, where ``dual-stage'' denotes schema guidance at two internal stages of the same end-to-end network: the proposal-generation stage and the decoder-refinement stage. At the proposal stage, a compact schema vector is inferred from encoder memory and used to guide initial pose prediction, ensuring that even the earliest object representations benefit from explicit structural supervision. At the decoder-refinement stage, schema is re-inferred from progressively refined query features and fed back into the pose head to condition iterative pose refinement. In this way, schema information is involved from coarse hypothesis generation to final pose decoding.

\paragraph{Schema-Guided Pose Head.}
Given the current object feature \(\mathbf{q}\), either from the encoder memory \(\mathbf{Q}_{enc}\) or decoder query features \(\mathbf{Q}_{dec}\), the Schema-Guided Pose Head first predicts the object category \(c\) and 2D bounding box \(\mathbf{b}\). The predicted box is fed back as spatial guidance to provide object-level localization cues for pose decoding. Based on \(\mathbf{q}\) and \(\mathbf{b}\), the head infers a compact schema vector, which is used as structural guidance for pose prediction. 

To support a unified prediction interface across all categories \(c \in \mathcal{C}\), we define a unified schema vector by concatenating category-wise schema slots:
\begin{equation}
    \mathbf{g} = [\mathbf{g}_1; \mathbf{g}_2; \dots; \mathbf{g}_n],
\end{equation}
where \(n=|\mathcal{C}|\) and \(\mathbf{g}_c\) denotes the compact schema vector for category \(c\). Since only a subset of schema dimensions is valid for each category, we apply a category-specific binary mask \(\mathbf{m}_c\) to obtain
\begin{equation}
    \tilde{\mathbf{g}} = \mathbf{g} \odot \mathbf{m}_c,
\end{equation}
where \(\odot\) denotes element-wise multiplication. This unified-but-masked design allows a single schema head to handle mixed-category prediction while preserving category-specific structural semantics.

Following \citet{simonelli2019disentangling}, we adopt a disentangled monocular pose parameterization. The head predicts four pose-related quantities:
\[
o = (o_x, o_y) \in \mathbb{R}^2, \quad 
z \in \mathbb{R}_+, \quad 
R \in SO(3), \quad 
s \in \mathbb{R}_+^3,
\]
corresponding to the projected object center, depth, rotation, and 3D size, respectively. 
In our implementation, the predicted bounding box \(\mathbf{b}\) is used as spatial guidance, and the effective schema vector \(\tilde{\mathbf{g}}\) is used as structural guidance for all four pose-related prediction branches.
Formally, the head output can be expressed as
\begin{equation}
    o, z, R, s = f_{\mathrm{SGPH}}\big([\mathbf{q}; \mathbf{b}; \tilde{\mathbf{g}}]\big),
\end{equation}
where \([\cdot\,;\cdot]\) denotes feature concatenation.

\paragraph{Pose Recovery.}
During inference, the 3D translation is recovered from the predicted projected center and depth using the camera intrinsic matrix \(K\):
\begin{equation}
t=zK^{-1}(o_x,o_y,1)^\top
\end{equation}
The final 9D pose prediction is then given by
\begin{equation}
\mathcal{P} = (R, t, s).
\end{equation}

\paragraph{Training.}
SchemaPose is trained end-to-end with unified supervision on detection, schema, and pose prediction. During training, the predictions of object queries are first matched to ground-truth instances by Hungarian matching under a one-to-one assignment. The matching cost is defined on object category and pose-related geometric quantities, including classification confidence, 2D bounding box, translation, and rotation, so that each positive query is assigned to a unique ground-truth instance while unmatched queries are treated as negatives. Based on the matched pairs, the overall objective is
\begin{equation}
\begin{aligned}
\mathcal{L}_{\text{total}}
=&\lambda_{\text{cls}}\mathcal{L}_{\text{cls}}
+\lambda_{\text{bbox}}\mathcal{L}_{\text{bbox}}
+\lambda_{\text{IoU}}\mathcal{L}_{\text{IoU}}
+\lambda_{\text{center2D}}\mathcal{L}_{\text{center2D}}
+\lambda_{\text{depth}}\mathcal{L}_{\text{depth}} \\
&+\lambda_{\text{rot}}\mathcal{L}_{\text{rot}}
+\lambda_{\text{scale}}\mathcal{L}_{\text{scale}}
+\lambda_{\text{schema}}^{\text{cont}}\mathcal{L}_{\text{schema}}^{\text{cont}}
+\lambda_{\text{schema}}^{\text{disc}}\mathcal{L}_{\text{schema}}^{\text{disc}} .
\end{aligned}
\end{equation}
We use focal loss~\cite{lin2017focal} for classification, \(\ell_1\) and GIoU losses~\cite{rezatofighi2019giou} for 2D boxes, \(\ell_1\) loss for projected center, \(\ell_2\) losses for depth and scale, geodesic distance for rotation, and \(\ell_1\) / sigmoid cross-entropy losses for continuous / discrete schema supervision.

For samples with schema annotations, both pose and schema losses are applied. For samples without schema annotations, the schema loss is masked out, while detection and pose losses are retained. This allows SchemaPose to use schema-supervised and pose-only data within the same end-to-end training objective.

\section{Experiments}

\subsection{Experimental Setup}

\paragraph{Benchmarks.}
We conduct experiments on the widely used category-level 9D pose estimation benchmarks CAMERA25 and REAL275~\cite{wang2019nocs}, following the standard six object categories: \textit{bottle}, \textit{bowl}, \textit{camera}, \textit{can}, \textit{laptop}, and \textit{mug}. CAMERA25 is a synthetic dataset containing 275K training images and 25K evaluation images, covering 1,269 object models from ShapeNetCore\cite{chang2015shapenet}. REAL275 is a real-world dataset consisting of 4.3K training images from 7 scenes and 2.75K test images from 6 scenes, with 6 object models in each category. We evaluate SchemaPose on the standard test splits of CAMERA25 and REAL275.

\paragraph{Training Data.}
Our training data are organized as a mixture of three sources. First, we use our schema-driven synthetic dataset, denoted as \textit{SchemaData}, generated by the pipeline introduced in Sec.~\ref{section schema}. \textit{SchemaData} provides scene-level synthetic RGB images with aligned pose labels and schema annotations. Unless otherwise specified, we instantiate 10K objects for each category and render 135K training images in total. Second, we use REAL275 training images with additional schema annotations provided by us, which helps schema-supervised learning transfer to real RGB observations. Third, we include CAMERA25 training images without schema annotations to further strengthen pose learning.

By default, the mixed training set consists of 135K images from \textit{SchemaData}, 135K randomly sampled images from the CAMERA25 training split, and the 4.3K REAL275 training images repeated 6 times. Schema supervision is applied to \textit{SchemaData} and REAL275 samples, and is masked out for CAMERA25 samples, while detection and pose supervision are retained for all training sources.

\paragraph{Evaluation Metrics.}
Following prior work~\cite{wang2019nocs,wei2024dmsr,lee2025yopo}, we report mean Average Precision (mAP) under 3D IoU thresholds and pose accuracy thresholds. Specifically, we use 3D IoU at 50\% and 75\%, and the standard pose metrics under the $10^\circ$ rotation error threshold, the 10 cm translation error threshold, and the combined $10^\circ$10 cm criterion.

\paragraph{Implementation Details.}
We use Swin-L~\cite{liu2021swin} as the backbone and initialize the detector from a DINO~\cite{zhang2022dino} model pretrained on COCO. We train the network with AdamW~\cite{loshchilov2017adamw} for 12 epochs on the mixed dataset and apply standard data augmentations, including resizing, color augmentation, random translation, and random flipping. All results are reported using a single model.

\subsection{Comparison with State-of-the-Art}

We compare SchemaPose with previous RGB-based category-level 9D pose estimation methods on CAMERA25 and REAL275. As shown in Table~\ref{tab:sota_main}, SchemaPose achieves the best results on all reported metrics across both benchmarks.

On CAMERA25, SchemaPose improves over the strongest prior method YOPO~\cite{lee2025yopo} by 5.8 IoU$_{50}$ and 4.4 under the $10^\circ$10cm metric. On REAL275, the gains are more pronounced: SchemaPose improves IoU$_{50}$ from 71.6 to 86.3 and $10^\circ$10cm from 52.8 to 68.4, corresponding to relative improvements of 20.5\% and 29.5\%, respectively. The consistent gains across both synthetic and real benchmarks validate SchemaPose as a strong approach for RGB-based category-level 9D pose estimation.

\begin{figure*}[t]
    \centering
    \includegraphics[trim=0 85 50 0,clip,width=\linewidth]{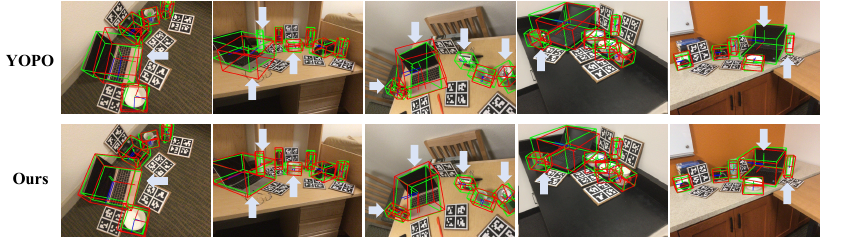}
    \vspace{-15pt}
    \caption{
    Qualitative comparison with YOPO on REAL275. 
    Green boxes indicate ground truth and red boxes indicate predictions. 
    Light-blue arrows mark instances where YOPO fails while SchemaPose produces more accurate pose estimates. 
    }
    \label{fig:qualitative}
\end{figure*}

Qualitative results are shown in Fig.~\ref{fig:qualitative}. Compared with YOPO, SchemaPose produces more accurate pose estimates in several challenging cases, especially under difficult viewpoints and cluttered real-world scenes. As highlighted by the light-blue arrows, YOPO often suffers from inaccurate pose prediction or even missed detections for small or heavily occluded instances, while SchemaPose can still localize these objects and estimate poses that better match the ground truth.  

We attribute the strong performance to the schema-guidance design of SchemaPose, whose motivation comes from the coupling between pose and object structure. The compact schema vector provides explicit cues about object extent, proportions, part layout, and symmetry, which help the pose head interpret monocular observations under intra-category structural variation. This allows the network to use a structured intermediate representation instead of relying solely on implicit query features for pose decoding, thereby making pose prediction more stable and better aligned with category-level structural variation.

\begin{table*}[t]
\centering
\small
\caption{Comparison with state-of-the-art RGB-based methods on CAMERA25 and REAL275. The best result in each column is shown in \textbf{bold}.}
\label{tab:sota_main}
\vspace{4pt}
\resizebox{\textwidth}{!}{
\begin{tabular}{l ccccc ccccc}
\toprule
\multirow{2}{*}{Method} 
    & \multicolumn{5}{c}{CAMERA25} 
    & \multicolumn{5}{c}{REAL275} \\
\cmidrule(lr){2-6} \cmidrule(lr){7-11}
    & IoU$_{50}$ & IoU$_{75}$ & 10cm & $10^\circ$ & $10^\circ$10cm
    & IoU$_{50}$ & IoU$_{75}$ & 10cm & $10^\circ$ & $10^\circ$10cm \\
\midrule
MSOS~\cite{lee2021msos}                  & 32.4 & 5.1 & 29.7 & 60.8 & 19.2 & 23.4 & 3.0 & 39.5 & 29.2 & 9.6  \\
OLD-Net~\cite{fan2022oldnet}             & 32.1 & 5.4 & 30.1 & 74.0 & 23.4 & 25.4 & 1.9 & 38.9 & 37.0 & 9.8  \\
DMSR~\cite{wei2024dmsr}                  & 34.6 & 6.5 & 32.3 & 81.4 & 27.4 & 28.3 & 6.1 & 37.3 & 59.5 & 23.6  \\
LaPose~\cite{zhang2024lapose}            & --   & --  & --   & --   & --   & 17.5 & 2.6 & 44.4 & --   & 30.5  \\
GIVEPose~\cite{huang2025givepose}        & --   & --  & --   & --   & --   & 20.1 & --  & 45.9 & --   & 34.2  \\
MonoDiff9D~\cite{liu2025monodiff9d}      & 35.2 & 6.7 & 33.6 & 80.1 & 28.2 & 31.5 & 6.3 & 41.0 & 56.3 & 25.7  \\
DA-Pose~\cite{yang2025dapose}            & 41.4 & 6.1 & 40.5 & 60.8 & 24.6 & 28.1 & 3.6 & 45.8 & 27.5 & 13.4  \\
Fischer et al.~\cite{fischer2025unified} & --  & -- & -- & -- & -- & 17.5 & 2.6 & -- & -- & 29.4 \\
LanCOPE~\cite{yang2025lancope}           & --   & --  & --   & --   & -- & 35.4 & 6.6 & 46.5 & 58.7 & 30.0 \\
YOPO~\cite{lee2025yopo}                  & 46.6 & 11.8 & 43.1 & 88.7 & 38.7 & 71.6 & 16.4& 77.8 & 69.6 & 52.8  \\
\midrule
SchemaPose (Ours)                        & \textbf{52.4} & \textbf{13.0} & \textbf{47.9} & \textbf{89.0} & \textbf{43.1} & \textbf{86.3} & \textbf{24.6} & \textbf{91.1} & \textbf{75.4} & \textbf{68.4} \\
\bottomrule
\end{tabular}
}
\end{table*}

\subsection{Ablation Study}

We conduct ablation studies on the REAL275 dataset to examine the effectiveness of our design. 

\begin{table}[htbp]
\setlength{\belowcaptionskip}{2pt} 
\centering
\begin{minipage}[t]{0.49\linewidth}
\centering
\caption{Ablation on schema-guided mechanism.}
\label{tab:ablation_dual_stage}
\footnotesize
\setlength{\tabcolsep}{3pt}
\begin{tabular}{lccccc}
\toprule
Method & IoU$_{50}$ & IoU$_{75}$ & 10cm & $10^\circ$ & $10^\circ$10cm \\
\midrule
\rowcolor{gray!15}
Dual-stage               & \textbf{86.3} & 24.6 & \textbf{91.1} & \textbf{75.4} & \textbf{68.4} \\
Decoder-only             & 86.0 & \textbf{25.8} & 89.6 & 74.3 & 66.0 \\
Proposal-only            & 82.7 & 25.0 & 86.0 & 74.0 & 63.3 \\
Supervision-only         & 82.5 & 24.6 & 87.8 & 74.5 & 66.5 \\
w/o Schema               & 80.5 & 22.5 & 85.5 & 75.1 & 63.8 \\
\bottomrule
\end{tabular}
\end{minipage}\hfill
\begin{minipage}[t]{0.49\linewidth}
\centering
\caption{Ablation on schema representation.}
\label{tab:ablation_schema_repr}
\footnotesize
\setlength{\tabcolsep}{3pt}
\begin{tabular}{lccccc}
\toprule
Representation & IoU$_{50}$ & IoU$_{75}$ & 10cm & $10^\circ$ & $10^\circ$10cm \\
\midrule
\rowcolor{gray!15}
Cont. + Disc.      & \textbf{86.3} & \textbf{24.6} & \textbf{91.1} & 75.4 & \textbf{68.4} \\
Cont. only         & 84.0 & 23.9 & 87.0 & 77.2 & 67.3 \\
Disc. only         & 82.6 & 22.1 & 83.2 & \textbf{77.7} & 64.9 \\
Shuffled schema    & 80.0 & 21.8 & 86.6 & 72.8 & 63.1 \\
Learned latent     & 83.4 & 22.9 & 83.6 & 73.0 & 61.8 \\
\bottomrule
\end{tabular}
\end{minipage}
\end{table}

\paragraph{Ablation on Dual-Stage Schema-Guided Mechanism.}
Table~\ref{tab:ablation_dual_stage} evaluates different ways of using schema.
The w/o Schema variant keeps the same backbone, query architecture, training data, and pose heads as SchemaPose, but removes schema supervision and schema guidance. Supervision-only improves over w/o Schema, showing that schema prediction provides useful structural supervision. However, using schema as explicit structural guidance for pose estimation is more effective: decoder-only guidance brings larger gains on IoU$_{50}$ and IoU$_{75}$, indicating that schema is especially helpful during refinement. The dual-stage design achieves the best overall performance, demonstrating that combining proposal-stage and decoder-stage schema guidance provides the most balanced benefit.

\paragraph{Ablation on Schema Representation.}
Table~\ref{tab:ablation_schema_repr} studies whether the gain comes from the structured schema semantics. The full schema, which combines continuous geometric attributes and discrete structural choices, achieves the best overall performance, showing that the two types of factors provide complementary information. Removing either component consistently degrades results, indicating that both are necessary for capturing complete structural information. We further compare with two semantics-broken variants: \emph{shuffled schema}, where schema annotations are randomly mismatched with object instances during training, and \emph{learned latent}, where the schema branch is replaced by a same-dimensional latent vector without schema supervision. Both variants perform worse than the true schema, suggesting that the improvement comes from the meaningful structural information encoded by schema, not merely from additional parameters or conditioning dimensions.

\begin{table}[htbp]
\setlength{\belowcaptionskip}{2pt} 
\begin{minipage}[t]{0.49\linewidth}
\centering
\caption{Ablation on schema-guided branches.}
\label{tab:ablation_schema_branches}
\scriptsize
\setlength{\tabcolsep}{2pt}
\begin{tabular}{cccccccccc}
\toprule
Center & Z & Rot & Size & Mask & IoU$_{50}$ & IoU$_{75}$ & 10cm & $10^\circ$ & $10^\circ$10cm \\
\midrule
        &        &            & \checkmark & \checkmark & 83.8 & 24.0 & 87.5 & 73.5 & 64.3 \\
        &        & \checkmark &            & \checkmark & 85.8 & 24.5 & 88.4 & 72.8 & 64.7 \\
        &        & \checkmark & \checkmark & \checkmark & 85.1 & 24.7 & 89.1 & 75.0 & 67.1  \\
\checkmark &        & \checkmark & \checkmark & \checkmark & 85.7 & \textbf{25.0} & 90.6 & 75.2 & 68.0 \\
        \rowcolor{gray!15}
\checkmark & \checkmark & \checkmark & \checkmark & \checkmark & \textbf{86.3} & 24.6 & \textbf{91.1} & \textbf{75.4} & \textbf{68.4} \\
\checkmark & \checkmark & \checkmark & \checkmark &            & 81.6 & 23.9 & 87.5 & 74.8 & 65.7 \\
\bottomrule
\end{tabular}
\end{minipage}\hfill
\begin{minipage}[t]{0.49\linewidth}
\centering
\caption{Ablation on training data composition.}
\label{tab:ablation_data_composition}
\scriptsize
\setlength{\tabcolsep}{2.4pt}
\begin{tabular}{cccccccc}
\toprule
SD & C25 & R275 & IoU$_{50}$ & IoU$_{75}$ & 10cm & $10^\circ$ & $10^\circ$10cm \\
\midrule
27K  & 135K & 4.3K$\times$6 & 83.5 & 22.5 & 90.0 & 71.1 & 62.9 \\
135K & 27K  & 4.3K$\times$6 & 84.8 & 26.0 & 89.7 & 74.7 & 65.7 \\
\rowcolor{gray!15}
135K & 135K & 4.3K$\times$6 & \textbf{86.3} & 24.6 & \textbf{91.1} & \textbf{75.4} & \textbf{68.4} \\
135K$^\dagger$ & 135K & 4.3K$\times$6 & 83.6 & \textbf{26.7} & 90.1 & 75.1 & 67.6 \\
270K & 135K & 4.3K$\times$6 & 84.4 & 24.0 & 88.6 & 74.8 & 66.0 \\
\bottomrule
\end{tabular}
{\scriptsize $^\dagger$Uses a reduced \textit{SchemaData} pool (200 instances per category).}
\end{minipage}
\vspace{-6pt}
\end{table}

\paragraph{Ablation on Schema-Guided Branches.}
Table~\ref{tab:ablation_schema_branches} studies how schema guidance affects different pose-related branches. Applying schema guidance only to size or rotation already improves performance, indicating that structural cues are useful for pose decoding. Guiding both rotation and size brings stronger results, while further extending schema guidance to the projected center and depth branches achieves the best overall performance on IoU$_{50}$, 10cm, $10^\circ$, and the combined $10^\circ$10cm metric. Removing the category-specific mask leads to a clear drop, showing that selecting valid schema slots for each category is important.

\paragraph{Ablation on Training Data Composition.}
Table~\ref{tab:ablation_data_composition} studies different training data compositions, where SD, C25, and R275 denote \textit{SchemaData}, CAMERA25, and REAL275, respectively. With the amount of REAL275 fixed, the default setting using 135K SD and 135K C25 achieves the best overall performance. Reducing SD to 27K or using the reduced-instance SD pool in 135K$^\dagger$ degrades performance, indicating that both the amount and structural diversity of schema-annotated data are important. Reducing C25 also hurts results, showing that pose-only CAMERA25 data still provides useful training signals. Increasing SD to 270K brings no further gain, suggesting that a balanced mixture of schema-annotated data, pose-only synthetic data, and real data is more effective than simply adding more schema data.

\section{Conclusion}

We presented SchemaPose, a schema-guided framework for RGB-based category-level 9D pose estimation. The core idea is to introduce a parametric category schema as an explicit structural variable that captures category-level commonality and instance-specific variation, and to use the inferred schema to guide pose prediction in a query-based network. To support schema learning, we further constructed a schema-driven data generation pipeline that provides aligned pose and schema annotations. Experiments on CAMERA25 and REAL275 show that SchemaPose achieves state-of-the-art performance among RGB-based methods, and ablation studies verify the effectiveness of dual-stage schema guidance, category-aware schema masking, structured schema representation, and balanced mixed training. These results suggest that explicit structural modeling is a promising direction for reducing monocular ambiguity and improving category-level pose estimation under large intra-category variation.

\begin{ack}


This work was supported by Fundamental and Interdisciplinary Disciplines Breakthrough Plan of the Ministry of Education of China, Science and Technology Major Project of Jiangsu Province (No. BG2025035), Shanghai Artificial Intelligence Laboratory, XPLORER PRIZE grants.
\end{ack}

\bibliographystyle{unsrtnat}
\bibliography{references}

@inproceedings{wang2019nocs,
  title={Normalized object coordinate space for category-level 6d object pose and size estimation},
  author={Wang, He and Sridhar, Srinath and Huang, Jingwei and Valentin, Julien and Song, Shuran and Guibas, Leonidas J},
  booktitle={Proceedings of the IEEE/CVF conference on computer vision and pattern recognition},
  pages={2642--2651},
  year={2019}
}

@article{lee2025yopo,
  title={You Only Pose Once: A Minimalist's Detection Transformer for Monocular RGB Category-level 9D Multi-Object Pose Estimation},
  author={Lee, Hakjin and Seo, Junghoon and Sim, Jaehoon},
  journal={arXiv preprint arXiv:2508.14965},
  year={2025}
}

@inproceedings{fischer2025unified,
  title={Unified Category-Level Object Detection and Pose Estimation from RGB Images using 3D Prototypes},
  author={Fischer, Tom and Zhang, Xiaojie and Ilg, Eddy},
  booktitle={Proceedings of the IEEE/CVF International Conference on Computer Vision},
  pages={9790--9800},
  year={2025}
}

@inproceedings{huang2025givepose,
  title={Givepose: Gradual intra-class variation elimination for rgb-based category-level object pose estimation},
  author={Huang, Ziqin and Wang, Gu and Zhang, Chenyangguang and Zhang, Ruida and Li, Xiu and Ji, Xiangyang},
  booktitle={Proceedings of the Computer Vision and Pattern Recognition Conference},
  pages={22055--22066},
  year={2025}
}

@inproceedings{zhang2024lapose,
  title={Lapose: Laplacian mixture shape modeling for rgb-based category-level object pose estimation},
  author={Zhang, Ruida and Huang, Ziqin and Wang, Gu and Zhang, Chenyangguang and Di, Yan and Zuo, Xingxing and Tang, Jiwen and Ji, Xiangyang},
  booktitle={European Conference on Computer Vision},
  pages={467--484},
  year={2024},
  organization={Springer}
}

@inproceedings{wei2024dmsr,
  title={Rgb-based category-level object pose estimation via decoupled metric scale recovery},
  author={Wei, Jiaxin and Song, Xibin and Liu, Weizhe and Kneip, Laurent and Li, Hongdong and Ji, Pan},
  booktitle={2024 IEEE International Conference on Robotics and Automation (ICRA)},
  pages={2036--2042},
  year={2024},
  organization={IEEE}
}

@article{lee2021msos,
  title={Category-level metric scale object shape and pose estimation},
  author={Lee, Taeyeop and Lee, Byeong-Uk and Kim, Myungchul and Kweon, In So},
  journal={IEEE Robotics and Automation Letters},
  volume={6},
  number={4},
  pages={8575--8582},
  year={2021},
  publisher={IEEE}
}

@inproceedings{liu2025monodiff9d,
  title={Monodiff9d: Monocular category-level 9d object pose estimation via diffusion model},
  author={Liu, Jian and Sun, Wei and Yang, Hui and Zheng, Jin and Geng, Zichen and Rahmani, Hossein and Mian, Ajmal},
  booktitle={2025 IEEE International Conference on Robotics and Automation (ICRA)},
  pages={8687--8694},
  year={2025},
  organization={IEEE}
}

@inproceedings{fan2022oldnet,
  title={Object level depth reconstruction for category level 6d object pose estimation from monocular rgb image},
  author={Fan, Zhaoxin and Song, Zhenbo and Xu, Jian and Wang, Zhicheng and Wu, Kejian and Liu, Hongyan and He, Jun},
  booktitle={European Conference on Computer Vision},
  pages={220--236},
  year={2022},
  organization={Springer}
}

@article{yang2025lancope,
  title={Lancope: Language-guided category-level object pose estimation from a single rgb image},
  author={Yang, Hui and Sun, Wei and Liu, Jian and Zheng, Jin and Dai, Zhenqi and Mian, Ajmal},
  journal={IEEE Robotics and Automation Letters},
  year={2025},
  publisher={IEEE}
}

@inproceedings{wan2023socs,
  title={Socs: Semantically-aware object coordinate space for category-level 6d object pose estimation under large shape variations},
  author={Wan, Boyan and Shi, Yifei and Xu, Kai},
  booktitle={Proceedings of the IEEE/CVF international conference on computer vision},
  pages={14065--14074},
  year={2023}
}

@inproceedings{ikeda2024diffusionnocs,
  title={Diffusionnocs: Managing symmetry and uncertainty in sim2real multi-modal category-level pose estimation},
  author={Ikeda, Takuya and Zakharov, Sergey and Ko, Tianyi and Irshad, Muhammad Zubair and Lee, Robert and Liu, Katherine and Ambrus, Rares and Nishiwaki, Koichi},
  booktitle={2024 IEEE/RSJ International Conference on Intelligent Robots and Systems (IROS)},
  pages={7406--7413},
  year={2024},
  organization={IEEE}
}

@inproceedings{he2017maskrcnn,
  title={Mask r-cnn},
  author={He, Kaiming and Gkioxari, Georgia and Doll{\'a}r, Piotr and Girshick, Ross},
  booktitle={Proceedings of the IEEE international conference on computer vision},
  pages={2961--2969},
  year={2017}
}

@inproceedings{simonelli2019disentangling,
  title={Disentangling monocular 3d object detection},
  author={Simonelli, Andrea and Bulo, Samuel Rota and Porzi, Lorenzo and L{\'o}pez-Antequera, Manuel and Kontschieder, Peter},
  booktitle={Proceedings of the IEEE/CVF international conference on computer vision},
  pages={1991--1999},
  year={2019}
}

@ARTICLE{yang2025dapose,
  author={Yang, Hui and Sun, Wei and Liu, Jian and Zheng, Jin and Zeng, Zhiwen and Mian, Ajmal},
  journal={IEEE Robotics and Automation Letters}, 
  title={RGB-Based Category-Level Object Pose Estimation via Depth Recovery and Adaptive Refinement}, 
  year={2025},
  volume={10},
  number={6},
  pages={5377-5384},
  doi={10.1109/LRA.2025.3559841}}

@inproceedings{fu20226d,
  title={6d robotic assembly based on rgb-only object pose estimation},
  author={Fu, Bowen and Leong, Sek Kun and Lian, Xiaocong and Ji, Xiangyang},
  booktitle={2022 IEEE/RSJ International Conference on Intelligent Robots and Systems (IROS)},
  pages={4736--4742},
  year={2022},
  organization={IEEE}
}

@article{tremblay2018deep,
  title={Deep object pose estimation for semantic robotic grasping of household objects},
  author={Tremblay, Jonathan and To, Thang and Sundaralingam, Balakumar and Xiang, Yu and Fox, Dieter and Birchfield, Stan},
  journal={arXiv preprint arXiv:1809.10790},
  year={2018}
}

@article{xiang2017posecnn,
  title={Posecnn: A convolutional neural network for 6d object pose estimation in cluttered scenes},
  author={Xiang, Yu and Schmidt, Tanner and Narayanan, Venkatraman and Fox, Dieter},
  journal={arXiv preprint arXiv:1711.00199},
  year={2017}
}

@article{lepetit2009epnp,
  title={EP n P: An accurate O (n) solution to the P n P problem},
  author={Lepetit, Vincent and Moreno-Noguer, Francesc and Fua, Pascal},
  journal={International journal of computer vision},
  volume={81},
  number={2},
  pages={155--166},
  year={2009},
  publisher={Springer}
}

@inproceedings{savva2019habitat,
  title={Habitat: A platform for embodied ai research},
  author={Savva, Manolis and Kadian, Abhishek and Maksymets, Oleksandr and Zhao, Yili and Wijmans, Erik and Jain, Bhavana and Straub, Julian and Liu, Jia and Koltun, Vladlen and Malik, Jitendra and others},
  booktitle={Proceedings of the IEEE/CVF international conference on computer vision},
  pages={9339--9347},
  year={2019}
}

@inproceedings{xiang2020sapien,
  title={Sapien: A simulated part-based interactive environment},
  author={Xiang, Fanbo and Qin, Yuzhe and Mo, Kaichun and Xia, Yikuan and Zhu, Hao and Liu, Fangchen and Liu, Minghua and Jiang, Hanxiao and Yuan, Yifu and Wang, He and others},
  booktitle={Proceedings of the IEEE/CVF conference on computer vision and pattern recognition},
  pages={11097--11107},
  year={2020}
}

@article{umeyama2002least,
  title={Least-squares estimation of transformation parameters between two point patterns},
  author={Umeyama, Shinji},
  journal={IEEE Transactions on pattern analysis and machine intelligence},
  volume={13},
  number={4},
  pages={376--380},
  year={2002},
  publisher={IEEE}
}

@article{marchand2015pose,
  title={Pose estimation for augmented reality: a hands-on survey},
  author={Marchand, Eric and Uchiyama, Hideaki and Spindler, Fabien},
  journal={IEEE transactions on visualization and computer graphics},
  volume={22},
  number={12},
  pages={2633--2651},
  year={2015},
  publisher={IEEE}
}

@article{tang20193d,
  title={3D mapping and 6D pose computation for real time augmented reality on cylindrical objects},
  author={Tang, Fulin and Wu, Yihong and Hou, Xiaohui and Ling, Haibin},
  journal={IEEE Transactions on Circuits and Systems for Video Technology},
  volume={30},
  number={9},
  pages={2887--2899},
  year={2019},
  publisher={IEEE}
}

@article{biederman1987recognition,
  title={Recognition-by-components: a theory of human image understanding.},
  author={Biederman, Irving},
  journal={Psychological review},
  volume={94},
  number={2},
  pages={115},
  year={1987},
  publisher={American Psychological Association}
}

@book{bartlett1995remembering,
  title={Remembering: A study in experimental and social psychology},
  author={Bartlett, Frederic Charles},
  year={1995},
  publisher={Cambridge university press}
}

@incollection{rumelhart2017schemata,
  title={Schemata: The building blocks of cognition},
  author={Rumelhart, David E},
  booktitle={Theoretical issues in reading comprehension},
  pages={33--58},
  year={2017},
  publisher={Routledge}
}

@article{zhang2022dino,
  title={Dino: Detr with improved denoising anchor boxes for end-to-end object detection},
  author={Zhang, Hao and Li, Feng and Liu, Shilong and Zhang, Lei and Su, Hang and Zhu, Jun and Ni, Lionel M and Shum, Heung-Yeung},
  journal={arXiv preprint arXiv:2203.03605},
  year={2022}
}

@inproceedings{lin2017focal,
  title={Focal loss for dense object detection},
  author={Lin, Tsung-Yi and Goyal, Priya and Girshick, Ross and He, Kaiming and Doll{\'a}r, Piotr},
  booktitle={Proceedings of the IEEE international conference on computer vision},
  pages={2980--2988},
  year={2017}
}

@inproceedings{rezatofighi2019giou,
  title={Generalized intersection over union: A metric and a loss for bounding box regression},
  author={Rezatofighi, Hamid and Tsoi, Nathan and Gwak, JunYoung and Sadeghian, Amir and Reid, Ian and Savarese, Silvio},
  booktitle={Proceedings of the IEEE/CVF conference on computer vision and pattern recognition},
  pages={658--666},
  year={2019}
}

@article{loshchilov2017adamw,
  title={Decoupled weight decay regularization},
  author={Loshchilov, Ilya and Hutter, Frank},
  journal={arXiv preprint arXiv:1711.05101},
  year={2017}
}

@inproceedings{liu2021swin,
  title={Swin transformer: Hierarchical vision transformer using shifted windows},
  author={Liu, Ze and Lin, Yutong and Cao, Yue and Hu, Han and Wei, Yixuan and Zhang, Zheng and Lin, Stephen and Guo, Baining},
  booktitle={Proceedings of the IEEE/CVF international conference on computer vision},
  pages={10012--10022},
  year={2021}
}

@article{bproc,
  author       = {Maximilian Denninger and
                  Martin Sundermeyer and
                  Dominik Winkelbauer and
                  Youssef Zidan and
                  Dmitry Olefir and
                  Mohamad Elbadrawy and
                  Ahsan Lodhi and
                  Harinandan Katam},
  title        = {BlenderProc},
  journal      = {CoRR},
  volume       = {abs/1911.01911},
  year         = {2019},
  url          = {http://arxiv.org/abs/1911.01911},
  eprinttype   = {arXiv},
  eprint       = {1911.01911},
  bibsource    = {dblp computer science bibliography, https://dblp.org}
}

@article{chang2015shapenet,
  title={Shapenet: An information-rich 3d model repository},
  author={Chang, Angel X and Funkhouser, Thomas and Guibas, Leonidas and Hanrahan, Pat and Huang, Qixing and Li, Zimo and Savarese, Silvio and Savva, Manolis and Song, Shuran and Su, Hao and others},
  journal={arXiv preprint arXiv:1512.03012},
  year={2015}
}

@inproceedings{zhou2019continuity,
  title={On the continuity of rotation representations in neural networks},
  author={Zhou, Yi and Barnes, Connelly and Lu, Jingwan and Yang, Jimei and Li, Hao},
  booktitle={Proceedings of the IEEE/CVF conference on computer vision and pattern recognition},
  pages={5745--5753},
  year={2019}
}

@inproceedings{wang2019densefusion,
  title={Densefusion: 6d object pose estimation by iterative dense fusion},
  author={Wang, Chen and Xu, Danfei and Zhu, Yuke and Mart{\'\i}n-Mart{\'\i}n, Roberto and Lu, Cewu and Fei-Fei, Li and Savarese, Silvio},
  booktitle={Proceedings of the IEEE/CVF conference on computer vision and pattern recognition},
  pages={3343--3352},
  year={2019}
}

@inproceedings{kato1999marker,
  title={Marker tracking and hmd calibration for a video-based augmented reality conferencing system},
  author={Kato, Hirokazu and Billinghurst, Mark},
  booktitle={Proceedings 2nd IEEE and ACM International Workshop on Augmented Reality (IWAR'99)},
  pages={85--94},
  year={1999},
  organization={IEEE}
}

@article{kolve2017ai2,
  title={Ai2-thor: An interactive 3d environment for visual ai},
  author={Kolve, Eric and Mottaghi, Roozbeh and Han, Winson and VanderBilt, Eli and Weihs, Luca and Herrasti, Alvaro and Deitke, Matt and Ehsani, Kiana and Gordon, Daniel and Zhu, Yuke and others},
  journal={arXiv preprint arXiv:1712.05474},
  year={2017}
}

\newpage
\appendix

\section*{Appendix Overview}
\addcontentsline{toc}{section}{Appendix Overview}

We provide comprehensive appendices for better understanding of our paper. The appendices are organized as follows:
\begin{itemize}
    \item Appendix~\ref{app:schema_implementation}: Implementation of \textbf{Parametric Category Schema}, including the primitive library, semantic part templates, category-level schema composition, and compact schema compression.
    \item Appendix~\ref{app:data_generation}: Details of \textbf{Schema-Driven Data Generation Pipeline}, including procedural object generation and compositional scene rendering.
    \item Appendix~\ref{appendix:implementation_details}: Architecture and training details of \textbf{SchemaPose framework}, including the Schema-Guided Pose Head, loss weights, matching costs, and training hyperparameters.
    \item Appendix~\ref{app:schema_discussion}: Discussions about the \textbf{role, expressiveness, and scalability} of the parametric category schema.
    \item Appendix~\ref{app:limitations}: Limitations and future work.
\end{itemize}

We also provide anonymized code, data, and qualitative videos in the supplementary material to support reproducibility. The code package includes the parametric schema templates, the schema-driven data generation pipeline, and the SchemaPose training and evaluation implementation. The released data include the generated SchemaData, procedurally instantiated object models, and schema annotations for REAL275. Together with the provided configurations, scripts, and annotations, these materials support reproduction of the main experiments and ablation studies reported in the paper.

\section{Implementation of Parametric Category Schema}
\label{app:schema_implementation}

A parametric category schema consists of three components: 
(1) a primitive library, 
(2) semantic part templates, and 
(3) category-level schema composition. 
The primitive library defines basic geometries such as \texttt{Cuboid}, \texttt{Cylinder}, and \texttt{Ring}. 
Semantic part templates instantiate and assemble these primitives according to continuous geometric parameters. 
Category-level schema composition selects active part templates according to discrete variables and assembles them into a complete object model. 
In addition, a compression function maps the full schema variable to the compact schema vector used for network learning.

\paragraph{Basic Geometry.}
We first show an example implementation of a basic geometry primitive. 
Each primitive is instantiated from a standard canonical template and then deformed by size parameters and transformed by local pose parameters.

\begin{verbatim}
class Cuboid():
    def __init__(self, size, position=[0,0,0], rotation=[0,0,0]):
        # Register parameters
        self.length = size[0]
        self.width  = size[1]
        self.height = size[2]
        self.position = position
        self.rotation = rotation

        # Standard cuboid template centered at origin
        self.vertices, self.faces = unit_cuboid()

        # Shape deformation
        self.vertices = self.vertices * [self.length,
                                         self.width,
                                         self.height]

        # Local pose transformation
        self.vertices = apply_pose(self.vertices,
                                   position=self.position,
                                   rotation=self.rotation)
\end{verbatim}

\paragraph{Part Template Definition.}
A semantic part template composes one or more primitives and defines their relative geometry. 
For example, a cylindrical mug body is instantiated from a ring-like wall and a bottom cylinder, while a trifold handle is instantiated from three cuboids. 
The input parameters determine both primitive dimensions and their relative placements.

\begin{verbatim}
class Cylindrical_Body():
    def __init__(self, Ri_t, Ro_t, Ri_b, Ro_b, h):
        # Register parameters
        self.Ri_t = Ri_t
        self.Ro_t = Ro_t
        self.Ri_b = Ri_b
        self.Ro_b = Ro_b
        self.h = h

        # Instantiate primitives
        self.wall = Ring(
            top_inner_radius    = Ri_t,
            top_outer_radius    = Ro_t,
            bottom_inner_radius = Ri_b,
            bottom_outer_radius = Ro_b,
            height = h
        )

        self.bottom = Cylinder(
            radius = Ro_b,
            height = bottom_thickness(Ro_b, Ri_b, h),
            position = [0, 0, -h / 2]
        )

        # Assemble primitives into a semantic part
        self.body = connect(self.wall, self.bottom)
\end{verbatim}

\begin{verbatim}
class Trifold_Handle():
    def __init__(self, lh_t, lh_b, lv, t, theta_t, theta_b):
        # Register parameters
        self.lh_t = lh_t
        self.lh_b = lh_b
        self.lv = lv
        self.t = t
        self.theta_t = theta_t
        self.theta_b = theta_b

        # Instantiate three cuboid bars
        self.top_bar = Cuboid(
            size = [lh_t, t, t],
            rotation = [0, 0, theta_t]
        )

        self.vertical_bar = Cuboid(
            size = [t, t, lv],
            position = handle_vertical_position(lh_t, lh_b, lv)
        )

        self.bottom_bar = Cuboid(
            size = [lh_b, t, t],
            rotation = [0, 0, theta_b]
        )

        # Assemble primitives into a handle
        self.handle = connect(self.top_bar,
                              self.vertical_bar,
                              self.bottom_bar)
\end{verbatim}

\paragraph{Category Schema Composition.}
At the category level, discrete variables select which part templates are active, while continuous variables provide the geometric parameters of those templates. 
The instantiation function \(\Phi_c\) corresponds to the following category-level construction.

\begin{verbatim}
class Mug_Schema():
    def __init__(self, d_c, x_c):
        # Discrete variables
        body_type   = d_c["body_type"]
        handle_type = d_c["handle_type"]

        # Continuous parameter groups
        body_params   = x_c["body"][body_type]
        handle_params = x_c["handle"][handle_type]

        # Instantiate active body template
        self.body = BodyTemplate(
            type   = body_type,
            params = body_params
        )

        # Instantiate optional handle template
        if handle_type != "none":
            self.handle = HandleTemplate(
                type   = handle_type,
                params = handle_params
            )
        else:
            self.handle = None

        # Assemble object model
        self.mesh = assemble(self.body, self.handle)

    def instantiate(self):
        # This implements Phi_c(d_c, x_c)
        return self.mesh
\end{verbatim}

\paragraph{Compact Schema Compression.}
The full schema variable contains all parameters needed for exact mesh instantiation. 
For network learning, we compress it into a compact schema vector by selecting pose-relevant factors, normalizing continuous parameters, and zero-padding inactive optional parts. 
This implements the category-specific compression function \(\Psi_c\).

\begin{verbatim}
def compress_mug_schema(d_c, x_c, ref_scale, valid_ranges):
    compact = []

    # Encode discrete template choices
    compact += one_hot(d_c["body_type"],
                       ["cylindrical", "prismatic", "multilevel"])
    compact += one_hot(d_c["handle_type"],
                       ["curved", "trifold", "none"])

    # Select and normalize pose-relevant body parameters
    body_type = d_c["body_type"]
    body = x_c["body"][body_type]
    compact += normalize([
        body["height"],
        body["outer_extent"],
        body["inner_extent"],
        body["thickness"]
    ], ref_scale, valid_ranges["body"])

    # Select and normalize pose-relevant handle parameters
    handle_type = d_c["handle_type"]
    if handle_type != "none":
        handle = x_c["handle"][handle_type]
        compact += normalize([
            handle["extent"],
            handle["offset"],
            handle["thickness"],
            handle["shape_ratio"]
        ], ref_scale, valid_ranges["handle"])
    else:
        compact += zeros(NUM_HANDLE_COMPACT_DIMS)

    return compact
\end{verbatim}

\section{Details of Schema-Driven Data Generation}
\label{app:data_generation}

This section details how we construct SchemaData. The pipeline has two stages: 
(1) procedural object generation from parametric category schemas, and 
(2) compositional scene rendering with BlenderProc\cite{bproc}.

\subsection{Procedural Object Generation}
\label{app:procedural_object_generation}

For each category \(c\), we generate object instances from the full schema variable
\[
    z_c=(d_c,x_c),
\]
where \(d_c\) specifies the discrete template choices and \(x_c\) specifies the continuous geometric parameters. Given \(z_c\), the instantiation function \(\Phi_c\) constructs an object mesh
\[
    M_c=\Phi_c(d_c,x_c).
\]
During generation, we save both the full schema variable \(z_c\) for reproducibility and the compact schema vector
\[
    \mathbf g_c=\Psi_c(z_c)
\]
as the schema annotation used for network supervision.

\paragraph{Discrete template choice sampling.}
We first sample the discrete variables \(d_c\) from the valid template choices of each category. These variables determine which semantic part templates are instantiated. For example, in the mug category, we sample the body type from cylindrical, prismatic, or multilevel templates, and sample the handle type from curved, trifold, or none. Similar optional or alternative templates are used for other categories, such as the lid of bottles, pull ring of cans, base of bowls, connector of laptops, and accessory parts of cameras. If a template is set to \texttt{none}, the corresponding part is not instantiated and its compact schema slots are zero-padded.

\paragraph{Continuous parameter sampling.}
After the active templates are selected, we sample their continuous parameters from category-specific valid intervals. These parameters control object dimensions, proportions, thicknesses, offsets, rotations, and part attachment geometry. For a parameter \(x_j\), we sample
\[
    x_j \sim \mathcal U(l_j,u_j),
\]
where \(l_j\) and \(u_j\) are defined in the schema configuration. Some bounds are conditional on other sampled parameters. For example, inner radii are constrained to be smaller than outer radii, wall thickness is constrained by the corresponding body extent, and optional-part offsets are constrained to keep the part attached to the supporting body.

\paragraph{Structure-preserving variation.}
To increase intra-category diversity, we vary schema parameters at three levels. First, we vary global object proportions, such as height, width, depth, and overall scale. Second, we vary part-level geometry, such as bottle neck size, bowl opening radius, laptop screen angle, camera lens length, or mug handle size. Third, we vary structural configuration through template choices, such as switching between different body or handle templates. These variations are applied within category-specific constraints so that the generated objects remain valid instances of the target category.

\paragraph{Validity checking.}
Each sampled schema variable is checked before export. We reject or resample objects with invalid geometry, including non-positive dimensions, invalid inner--outer relations, excessive thickness, degenerate levels, disconnected optional parts, severe primitive interpenetration, or floating components. This step prevents unrealistic meshes caused by extreme parameter combinations.

\paragraph{Mesh export and metadata.}
For each valid object, we export the mesh in the category canonical coordinate system. We also save its category label, full schema variable \(z_c\), compact schema vector \(\mathbf g_c\), object scale, and reference scale. The compact vector \(\mathbf g_c\) is computed by selecting pose-relevant schema factors, normalizing continuous parameters by category-specific reference scales or valid ranges, and zero-padding inactive optional-part slots.

\begin{figure*}[t]
    \centering
    \includegraphics[trim=50 10 50 0,clip,width=\linewidth]{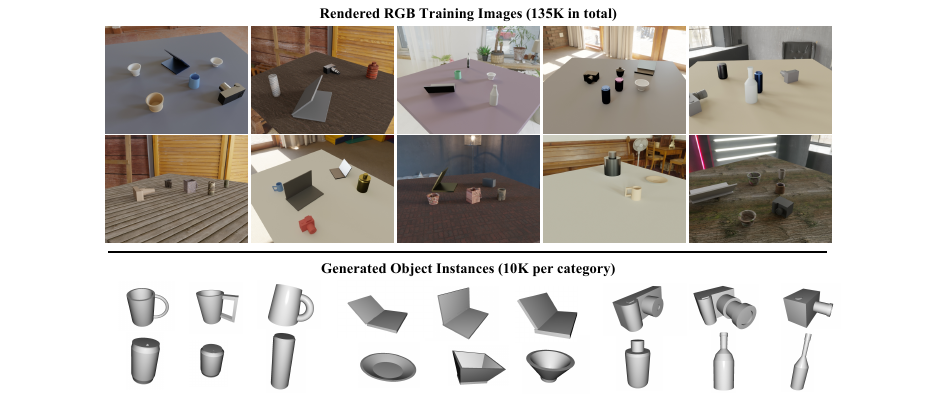}
    \caption{\textbf{Additional examples of \textit{SchemaData}.}
    The top rows show rendered RGB training images generated by compositional scene rendering with diverse object layouts, viewpoints, lighting conditions, materials, textures, and backgrounds. The bottom row shows schema-generated object instances from different categories. These objects are obtained by varying discrete template choices and continuous geometric parameters within category-consistent ranges. Each rendered instance is associated with aligned pose labels and a compact schema vector inherited from its generation parameters.}
    \label{fig:schemadata_examples}
\end{figure*}

\subsection{Compositional Scene Rendering}
\label{app:scene_rendering}

Given the generated object pool, we use BlenderProc to render RGB training images. The goal is to construct cluttered multi-object scenes that expose the network to diverse object poses, layouts, occlusions, and appearance conditions.

\paragraph{Scene initialization.}
Each training scene is initialized with a supporting surface, such as a table plane or a floor-like plane, together with a background environment and rendering assets. We randomize the supporting surface geometry and its material appearance so that the rendered scenes do not overfit to a fixed tabletop texture or color. The scene may also contain additional non-target objects or background clutter, depending on the rendering configuration.

\paragraph{Object subset sampling.}
For each image, we sample a variable number of generated objects from the object pool. The sampled objects may come from different categories, which produces mixed-category scenes. For each sampled object, we load both its mesh and its metadata, including category label, full schema variable, compact schema vector, and canonical size. This ensures that every rendered instance remains associated with its structural annotation throughout the rendering process.

\paragraph{Physics-aware object placement.}
After object sampling, we place the objects into the scene in a physically plausible manner. Concretely, each object is first assigned a random initial position above the supporting surface and a random initial orientation. We then use BlenderProc's rigid-body simulation to let the objects settle under gravity until a stable scene configuration is reached. This step automatically resolves many unrealistic placements and produces natural object contacts and occlusions. To avoid degenerate scenes, we reject configurations with severe object interpenetration, unstable placements, or an excessive number of objects falling outside the camera view.

\paragraph{Pose and scale variation.}
Before or during placement, each object is additionally assigned random scale and pose variation within category-consistent limits. The scale variation is applied to simulate intra-category size diversity, while the 3D rotation and translation are varied to cover a broad pose distribution. Since these transformations are applied to schema-generated objects whose canonical geometry is already known, the final rendered pose labels remain exactly aligned with the instantiated mesh.

\paragraph{Layout diversity and occlusion control.}
To improve robustness to cluttered scenes, we explicitly vary the number of objects, their relative distances, and their degree of overlap. This produces images ranging from isolated-object cases to heavily cluttered scenes with partial occlusion. During rendering, we keep only visible instances whose projected area exceeds a minimum threshold and whose truncation is not excessively large. This avoids adding nearly invisible objects as noisy supervision.

\paragraph{Camera sampling.}
We sample the camera pose around the scene by varying azimuth, elevation, and distance relative to the object layout region. The camera is oriented toward the scene center so that most sampled objects remain within view. For each rendered image, the intrinsic matrix \(K\) and the camera extrinsic parameters are recorded. This allows us to export pose labels directly in the camera coordinate system and to train the network under the same monocular RGB setting used at test time.

\paragraph{Appearance randomization.}
To reduce the domain gap between synthetic and real images, we randomize multiple rendering factors independently of the schema parameters. These factors include:
\begin{itemize}
    \item light source type, position, intensity, and color;
    \item material properties such as roughness, metallicness, and specularity;
    \item object colors and surface textures;
    \item background appearance and environment illumination.
\end{itemize}
This decouples structural variation from appearance variation, so that the generated dataset covers both shape diversity and rendering diversity.

\paragraph{Rendering outputs and annotation export.}
For each scene, BlenderProc renders an RGB image together with instance-level visibility information. Since the complete scene state is known, we export for every visible target object its category label, 2D bounding box, segmentation mask, 3D rotation, 3D translation, and 3D size. In addition, the compact schema vector \(\mathbf g_c\) is directly inherited from the object metadata generated in the previous stage. Therefore, the rendered training images provide aligned supervision at both the pose level and the schema level.

\paragraph{Filtering and quality control.}
After rendering, we apply a final filtering step to remove low-quality annotations. Specifically, an object instance is discarded if its visible area is too small, if its projected bounding box is below a minimum size, or if the visible portion is too heavily truncated. This filtering prevents extremely hard or ambiguous synthetic samples from dominating the supervision signal.

\subsection{REAL275 Schema Annotation}
\label{app:real275_schema_annotation}

In addition to schema annotations generated automatically for SchemaData, we also provide schema annotations for the REAL275 training split to support schema-supervised learning on real RGB images. The annotation is performed only on the REAL275 training objects and does not use any REAL275 test images, test object instances, or test annotations.

For each REAL275 training object, annotators assign the discrete template choices and continuous schema parameters according to the predefined category schema. The annotation follows the same schema definitions used for synthetic object generation. Specifically, annotators first select the part templates that best match the object structure, and then adjust the continuous parameters so that the object instantiated from the schema is as close as possible to the corresponding real object model. This process produces a compact schema vector through the same compression function \(\Psi_c\), which is used as schema supervision during training.

REAL275 contains six object categories, with three training object instances per category. In our annotation process, each object takes about 20 minutes on average. Since the annotation is performed at the object-instance level rather than the image level, the resulting schema label can be reused for all training images containing the same object. At inference time, SchemaPose does not require any schema annotation or object model; it predicts the schema vector directly from the input RGB image.

\section{Details of Architecture and Training Configuration}
\label{appendix:implementation_details}

This section provides additional implementation details for reproducing SchemaPose. Unless otherwise specified, all settings follow the configuration used for the main results.

\subsection{Network Architecture}
\label{appendix:network_architecture}

Following the design of DINO\cite{zhang2022dino}, SchemaPose is implemented with a two-stage query-based detector with box refinement. We use Swin-L\cite{liu2021swin} as the backbone and initialize the detector from a DINO model pretrained on COCO. The backbone uses an embedding dimension of 192, depths of \([2,2,18,2]\), attention heads of \([6,12,24,48]\), a window size of 12, and a drop-path rate of 0.2. Feature maps from four backbone stages with channel dimensions \([192,384,768,1536]\) are projected by a ChannelMapper neck to 256 channels and expanded to 5 feature levels.

The Transformer encoder and decoder both contain 6 layers. The hidden dimension is 256, the feed-forward dimension is 2048, and the decoder uses 8 attention heads. The sine positional encoding uses 128 positional features. At the proposal stage, encoder-side classification scores and boxes are predicted from encoder output features, and the top-100 proposals are selected by maximum classification score to initialize decoder reference queries. In parallel, SGPH is applied to encoder-side features to produce auxiliary detection, schema, and pose predictions for training. During inference, only the predictions from the last decoder layer are used as final outputs.

\subsection{Schema-Guided Pose Head}
\label{appendix:sgph_details}

The Schema-Guided Pose Head predicts object category, 2D bounding box, compact schema, projected center, depth, rotation, and 3D size. It is applied at both the proposal stage and the decoder stage. The prediction branches are not shared across decoder layers; each prediction layer has its own branch parameters following the standard DINO-style refinement design.

The classification branch is a linear layer from the 256-dimensional object feature to 6 category logits. The 2D box branch is a 2-layer MLP with ReLU activations and predicts a normalized box \((c_x,c_y,w,h)\). The compact schema branch takes the object feature concatenated with the predicted 2D box as input and predicts a 60-dimensional unified schema vector, consisting of 44 continuous dimensions and 16 discrete dimensions.

The compact schema vector is organized as category-wise slots. The continuous dimensions for \textit{bottle}, \textit{bowl}, \textit{camera}, \textit{can}, \textit{laptop}, and \textit{mug} are \((11,5,6,7,7,8)\), respectively. The discrete dimensions are \((3,4,2,2,1,4)\), respectively. During schema conditioning, we construct a category-specific binary mask from the predicted category and keep only the schema slots corresponding to that category. This masked schema vector is then used as the effective structural condition.

For pose prediction, the projected center, depth, rotation, and size branches are implemented as MLPs with two hidden layers. The predicted 2D box and the effective schema vector are concatenated to all four pose branches, serving as spatial and structural guidance, respectively. Therefore, all pose-related branches take a 320-dimensional input, formed by concatenating the 256-dimensional object feature, 4-dimensional 2D box, and 60-dimensional effective schema vector. Rotation is represented by the continuous 6D representation~\cite{zhou2019continuity}. We use class-wise rotation and size heads: the rotation branch outputs \(6 \times 6 = 36\) values, and the size branch outputs \(6 \times 3 = 18\) values. During inference, the rotation and size corresponding to the predicted category are selected.

\begin{table}[h]
\centering
\caption{Branch configuration of the Schema-Guided Pose Head.}
\label{tab:sgph_branch_config}
\small
\begin{tabular}{lcc}
\toprule
Branch & Input & Output dim. \\
\midrule
Classification & object feature & 6 \\
2D box & object feature & 4 \\
Schema & feature + box & 60 \\
Projected center & feature + box + schema & 2 \\
Depth & feature + box + schema & 1 \\
Rotation & feature + box + schema & 36 \\
Size & feature + box + schema & 18 \\
\bottomrule
\end{tabular}
\end{table}

\begin{table}[h]
\centering
\caption{Compact schema dimensions for each category.}
\label{tab:appendix_schema_dims}
\begin{tabular}{lccc}
\toprule
Category & Continuous dim. & Discrete dim. & Total dim. \\
\midrule
Bottle & 11 & 3 & 14 \\
Bowl & 5 & 4 & 9 \\
Camera & 6 & 2 & 8 \\
Can & 7 & 2 & 9 \\
Laptop & 7 & 1 & 8 \\
Mug & 8 & 4 & 12 \\
\midrule
Total & 44 & 16 & 60 \\
\bottomrule
\end{tabular}
\end{table}

\subsection{Loss Weights and Matching Costs}
\label{appendix:loss_matching_details}

SchemaPose is trained with detection, pose, and schema losses. For detection, we use focal loss for classification, \(\ell_1\) loss for 2D box regression, and GIoU loss for box overlap. For pose prediction, we use \(\ell_1\) loss for the projected center, \(\ell_2\) loss for depth, geodesic rotation loss for rotation, and \(\ell_2\) loss for 3D size. For schema prediction, we use \(\ell_1\) loss for continuous schema dimensions and sigmoid cross-entropy loss for discrete schema dimensions. Schema losses are applied only when schema annotations are available; otherwise, the schema loss weights are masked out for that sample.

The loss weights are listed in Table~\ref{tab:appendix_loss_weights}. We apply these losses to the final decoder layer, intermediate decoder layers, encoder-side auxiliary outputs, and denoising outputs. The final reported predictions are still taken only from the last decoder layer during inference.

\begin{table}[h]
\centering
\caption{Loss weights used for training SchemaPose.}
\label{tab:appendix_loss_weights}
\begin{tabular}{lc}
\toprule
Loss term & Weight \\
\midrule
Classification focal loss & 1.0 \\
BBox \(\ell_1\) loss & 5.0 \\
BBox GIoU loss & 2.0 \\
Projected center \(\ell_1\) loss & 5.0 \\
Depth \(\ell_2\) loss & 50.0 \\
Rotation geodesic loss & 5.0 \\
Size \(\ell_2\) loss & 50.0 \\
Continuous schema \(\ell_1\) loss & 0.4 \\
Discrete schema sigmoid CE loss & 0.4 \\
\bottomrule
\end{tabular}
\end{table}

For one-to-one assignment, we use Hungarian matching with classification, 2D box, GIoU, translation, and rotation costs. The matching costs are listed in Table~\ref{tab:appendix_matching_costs}. We do not use schema prediction in the matching cost. For rotation matching, we follow the standard NOCS symmetry setting and treat \textit{bottle}, \textit{bowl}, and \textit{can} as symmetric categories.

\begin{table}[h]
\centering
\caption{Hungarian matching cost weights.}
\label{tab:appendix_matching_costs}
\begin{tabular}{lc}
\toprule
Matching cost term & Weight \\
\midrule
Focal classification cost & 2.0 \\
BBox \(\ell_1\) cost & 5.0 \\
GIoU cost & 2.0 \\
Translation cost & 5.0 \\
Rotation cost & 2.0 \\
\bottomrule
\end{tabular}
\end{table}

\subsection{Training Hyperparameters}
\label{appendix:training_hyperparameters}

We train SchemaPose on 4 NVIDIA H200 GPUs with a per-GPU batch size of 8, resulting in a total batch size of 32. We use AdamW with a base learning rate of \(1\times10^{-4}\) and enable automatic learning-rate scaling with a reference batch size of 16. Therefore, the effective learning rate is scaled to \(2\times10^{-4}\). The weight decay is \(1\times10^{-4}\), and the backbone learning rate is multiplied by 0.1. We apply gradient clipping with a maximum norm of 0.1.

\subsection{Statistical Variation Across Random Seeds}

To assess the stability of SchemaPose, we repeat the main REAL275 experiment with five random seeds and report the mean and standard deviation. All runs use the same architecture, training data composition, and hyperparameters, and differ only in random initialization and data shuffling. As shown in Table~\ref{tab:seed_variation}, the performance variation is small, indicating that the reported improvement is stable across random seeds.

\begin{table}[h]
\centering
\caption{Statistical variation of SchemaPose on REAL275 across five random seeds. We report mean $\pm$ standard deviation.}
\label{tab:seed_variation}
\begin{tabular}{lccccc}
\toprule
Method & IoU$_{50}$ & IoU$_{75}$ & 10cm & $10^\circ$ & $10^\circ$10cm \\
\midrule
SchemaPose & 86.5$\pm$1.0 & 25.8$\pm$1.5 & 90.8$\pm$0.3 & 74.9$\pm$1.0 & 67.5$\pm$1.3 \\
\bottomrule
\end{tabular}
\end{table}

\section{Discussion on Parametric Category Schema}
\label{app:schema_discussion}

\paragraph{Schema as a Pose-Oriented Structural Prior.}
The parametric category schema provides an explicit structural context for category-level pose estimation. In the RGB-only setting, the same image evidence may correspond to different 3D interpretations because object extent, proportions, part layout, and symmetry vary across instances within the same category. Schema makes these factors explicit. Continuous schema parameters describe metric-relevant properties such as object size, aspect ratio, thickness, and part offsets, which help the network reason about 3D scale and translation under monocular ambiguity. Discrete schema variables describe structural choices such as part existence, template type, and coarse configuration, which provide cues for resolving orientation ambiguity caused by symmetry or anisotropic object layouts.

This structural information complements visual query features. Without schema, the network must encode category-level commonality, instance-specific shape variation, and pose cues in the same latent representation. This can make pose prediction sensitive to irrelevant appearance details or instance-specific geometry. By contrast, schema separates pose-relevant structure into an explicit intermediate variable. The pose head can then condition its prediction on both visual evidence and structured object information, making rotation, translation, and size estimation more aligned with the category-level 9D pose objective.

Importantly, schema is not a fixed prototype or a substitute for visual learning. It represents a parameterized family of valid category instances, while the network still learns to infer schema and pose from RGB observations. Its role is to organize the structural variation that affects pose prediction, so that the estimator does not need to recover all such factors only through unconstrained latent features.

\paragraph{Expressiveness from Primitive Composition.}
Parametric Category Schema can comprehensively describe the structural information of various objects encountered in daily life. This expressive power of schema comes from the composition of reusable geometric primitives and semantic part templates. Basic primitives, such as cuboids, cylinders, rings, and other simple parameterized shapes, describe common local geometry, while semantic templates organize these primitives into meaningful object parts. This design follows the observation that many man-made objects can be described as combinations of basic geometric components, where each primitive captures a family of shapes through its parameters and their composition yields diverse object structures\cite{biederman1987recognition,rumelhart2017schemata}. 
As a reference, PartNet-Mobility~\cite{xiang2020sapien} is a large-scale articulated object dataset collected from real-world manufacturer-level 3D models, covering substantial intra-category structural variation. For the six pose-estimation categories considered in this work, our parametric category schemas can represent nearly all corresponding instances from \citet{xiang2020sapien} by composing the predefined primitives and part templates with category-specific parameter ranges. This coverage suggests that the proposed schema representation is expressive enough to capture the major structural variations required for category-level pose estimation, while remaining compact and analytically controllable.

\paragraph{Scalability of Schema Construction.}
As mentioned above, the strong expressiveness of reusable geometric primitives and semantic part templates enables parametric schema to easily scale to novel object categories. Extending SchemaPose to a new category usually starts by identifying reusable components from the existing part templates, and then specifying category-specific composition rules and parameter ranges. For example, handle-like templates can be reused for mugs, drawers, or containers, while cylindrical body templates can support many container-like categories with different proportions and attachment rules. If the existing library is insufficient, a new template can be implemented by inheriting from existing part templates and invoking basic geometry templates. In our implementation, developing parametric schema for a novel category takes about 2 hours on average. Once implemented, the new templates are added back to the schema library and can be reused by later categories, making the construction process cumulative and increasingly efficient as the library grows, rather than a from-scratch design process for every category.

\paragraph{Consistency Between Generation and Learning.}
The same schema formulation is used both for object generation and for network learning. During data generation, the full schema variable specifies template choices, geometric parameters, and analytic relations needed to instantiate valid object meshes. During learning, a compact schema vector derived from the full schema variable provides pose-relevant structural supervision. This consistency is important: the structural variables used to generate synthetic object instances are the same variables that the network learns to infer and use for pose prediction. As a result, schema provides not only a generative specification for constructing diverse training data, but also an explicit intermediate representation that connects image observations with category-level pose estimation.

\section{Limitations and Future Work}
\label{app:limitations}

While SchemaPose demonstrates the effectiveness of introducing explicit schema information into RGB-based category-level pose estimation, several directions remain open for future exploration.

\paragraph{Automatic Schema Construction.}
In the current implementation, category schemas are designed manually based on geometric primitives, semantic part templates, and category-level structural analysis. This design makes the resulting schema interpretable and directly usable for both object generation and pose prediction. A promising future direction is to reduce the amount of manual design by developing semi-automatic or data-driven tools for schema construction. For example, candidate primitives, part templates, or parameter ranges could be proposed from collections of 3D models or generated instances, and then refined into compact schema definitions with minimal human intervention.

\paragraph{Expanding the Schema Library.}
The current schema library is constructed for the object categories studied in our experiments. Since schemas are built from reusable geometric primitives and semantic part templates, the library can be extended cumulatively as more categories are considered. Future work will expand the collection of primitives, templates, and category-specific composition rules, so that a broader range of object categories can share and reuse existing schema components. Such an expanding library would also make schema construction increasingly efficient, as newly defined templates can benefit later categories.

\paragraph{Improving Synthetic-to-Real Transfer.}
The schema-driven data generation pipeline provides accurate schema and pose annotations by construction, but the visual distribution of synthetic data can still differ from real RGB observations. Although mixed training and rendering randomization reduce this gap, future work can further improve synthetic-to-real transfer through stronger domain randomization, more realistic material and lighting simulation, or image-level adaptation. This direction is particularly important for making schema-supervised learning more effective in real-world deployment.

\paragraph{Temporal and Multi-View Extensions.}
SchemaPose estimates category-level 9D pose from a single RGB image. In many practical scenarios, however, multiple frames or multiple views are available. Extending schema-guided pose estimation to temporal or multi-view settings could provide additional constraints for resolving ambiguity in monocular observations. Since schema represents stable object structure, it can naturally serve as a consistency variable across frames or views, enabling more robust pose estimation under occlusion, viewpoint change, and clutter.

Overall, these directions suggest that schema-guided pose estimation is not limited to the current implementation, but can be further developed through more automatic schema construction, richer schema libraries, improved synthetic-to-real transfer, and stronger temporal or multi-view reasoning.

\section{Large Language Model Usage Statement}
We promise that LLMs are used only for spelling, grammar checking and phrasing to improve the paper's clarity for easier understanding. All research ideas, methods, experiments, and conclusions are the sole work of the authors.


\newpage
\section*{NeurIPS Paper Checklist}

\begin{enumerate}

\item {\bf Claims}
    \item[] Question: Do the main claims made in the abstract and introduction accurately reflect the paper's contributions and scope?
    \item[] Answer: \answerYes{} 
    \item[] Justification: The abstract and introduction state the main contributions of SchemaPose, including the schema-guided framework, parametric category schema, schema-driven data generation, and empirical validation. The claims are supported by the method and experimental sections.
    \item[] Guidelines:
    \begin{itemize}
        \item The answer \answerNA{} means that the abstract and introduction do not include the claims made in the paper.
        \item The abstract and/or introduction should clearly state the claims made, including the contributions made in the paper and important assumptions and limitations. A \answerNo{} or \answerNA{} answer to this question will not be perceived well by the reviewers. 
        \item The claims made should match theoretical and experimental results, and reflect how much the results can be expected to generalize to other settings. 
        \item It is fine to include aspirational goals as motivation as long as it is clear that these goals are not attained by the paper. 
    \end{itemize}

\item {\bf Limitations}
    \item[] Question: Does the paper discuss the limitations of the work performed by the authors?
    \item[] Answer: \answerYes{} 
    \item[] Justification: The paper discusses the scope and limitations of schema construction, including its reliance on parametric templates and its applicability to object categories whose structural variation can be reasonably captured by such templates.
    \item[] Guidelines:
    \begin{itemize}
        \item The answer \answerNA{} means that the paper has no limitation while the answer \answerNo{} means that the paper has limitations, but those are not discussed in the paper. 
        \item The authors are encouraged to create a separate ``Limitations'' section in their paper.
        \item The paper should point out any strong assumptions and how robust the results are to violations of these assumptions (e.g., independence assumptions, noiseless settings, model well-specification, asymptotic approximations only holding locally). The authors should reflect on how these assumptions might be violated in practice and what the implications would be.
        \item The authors should reflect on the scope of the claims made, e.g., if the approach was only tested on a few datasets or with a few runs. In general, empirical results often depend on implicit assumptions, which should be articulated.
        \item The authors should reflect on the factors that influence the performance of the approach. For example, a facial recognition algorithm may perform poorly when image resolution is low or images are taken in low lighting. Or a speech-to-text system might not be used reliably to provide closed captions for online lectures because it fails to handle technical jargon.
        \item The authors should discuss the computational efficiency of the proposed algorithms and how they scale with dataset size.
        \item If applicable, the authors should discuss possible limitations of their approach to address problems of privacy and fairness.
        \item While the authors might fear that complete honesty about limitations might be used by reviewers as grounds for rejection, a worse outcome might be that reviewers discover limitations that aren't acknowledged in the paper. The authors should use their best judgment and recognize that individual actions in favor of transparency play an important role in developing norms that preserve the integrity of the community. Reviewers will be specifically instructed to not penalize honesty concerning limitations.
    \end{itemize}

\item {\bf Theory assumptions and proofs}
    \item[] Question: For each theoretical result, does the paper provide the full set of assumptions and a complete (and correct) proof?
    \item[] Answer:\answerNA{} 
    \item[] Justification:  The paper does not include theoretical results.
    \item[] Guidelines:
    \begin{itemize}
        \item The answer \answerNA{} means that the paper does not include theoretical results. 
        \item All the theorems, formulas, and proofs in the paper should be numbered and cross-referenced.
        \item All assumptions should be clearly stated or referenced in the statement of any theorems.
        \item The proofs can either appear in the main paper or the supplemental material, but if they appear in the supplemental material, the authors are encouraged to provide a short proof sketch to provide intuition. 
        \item Inversely, any informal proof provided in the core of the paper should be complemented by formal proofs provided in appendix or supplemental material.
        \item Theorems and Lemmas that the proof relies upon should be properly referenced. 
    \end{itemize}

    \item {\bf Experimental result reproducibility}
    \item[] Question: Does the paper fully disclose all the information needed to reproduce the main experimental results of the paper to the extent that it affects the main claims and/or conclusions of the paper (regardless of whether the code and data are provided or not)?
    \item[] Answer: \answerYes{} 
    \item[] Justification: The paper describes the model architecture, schema representation, data generation pipeline, training objectives, datasets, evaluation metrics, and ablation settings needed to reproduce the main experimental results.
    \item[] Guidelines:
    \begin{itemize}
        \item The answer \answerNA{} means that the paper does not include experiments.
        \item If the paper includes experiments, a \answerNo{} answer to this question will not be perceived well by the reviewers: Making the paper reproducible is important, regardless of whether the code and data are provided or not.
        \item If the contribution is a dataset and\slash or model, the authors should describe the steps taken to make their results reproducible or verifiable. 
        \item Depending on the contribution, reproducibility can be accomplished in various ways. For example, if the contribution is a novel architecture, describing the architecture fully might suffice, or if the contribution is a specific model and empirical evaluation, it may be necessary to either make it possible for others to replicate the model with the same dataset, or provide access to the model. In general. releasing code and data is often one good way to accomplish this, but reproducibility can also be provided via detailed instructions for how to replicate the results, access to a hosted model (e.g., in the case of a large language model), releasing of a model checkpoint, or other means that are appropriate to the research performed.
        \item While NeurIPS does not require releasing code, the conference does require all submissions to provide some reasonable avenue for reproducibility, which may depend on the nature of the contribution. For example
        \begin{enumerate}
            \item If the contribution is primarily a new algorithm, the paper should make it clear how to reproduce that algorithm.
            \item If the contribution is primarily a new model architecture, the paper should describe the architecture clearly and fully.
            \item If the contribution is a new model (e.g., a large language model), then there should either be a way to access this model for reproducing the results or a way to reproduce the model (e.g., with an open-source dataset or instructions for how to construct the dataset).
            \item We recognize that reproducibility may be tricky in some cases, in which case authors are welcome to describe the particular way they provide for reproducibility. In the case of closed-source models, it may be that access to the model is limited in some way (e.g., to registered users), but it should be possible for other researchers to have some path to reproducing or verifying the results.
        \end{enumerate}
    \end{itemize}

\item {\bf Open access to data and code}
    \item[] Question: Does the paper provide open access to the data and code, with sufficient instructions to faithfully reproduce the main experimental results, as described in supplemental material?
    \item[] Answer: \answerYes{} 
    \item[] Justification:  We provide anonymized code, configuration files, data generation scripts, and instructions to reproduce the main experiments in the supplementary material. The generated synthetic dataset is provided through an anonymous URL due to its size.
    \item[] Guidelines:
    \begin{itemize}
        \item The answer \answerNA{} means that paper does not include experiments requiring code.
        \item Please see the NeurIPS code and data submission guidelines (\url{https://neurips.cc/public/guides/CodeSubmissionPolicy}) for more details.
        \item While we encourage the release of code and data, we understand that this might not be possible, so \answerNo{} is an acceptable answer. Papers cannot be rejected simply for not including code, unless this is central to the contribution (e.g., for a new open-source benchmark).
        \item The instructions should contain the exact command and environment needed to run to reproduce the results. See the NeurIPS code and data submission guidelines (\url{https://neurips.cc/public/guides/CodeSubmissionPolicy}) for more details.
        \item The authors should provide instructions on data access and preparation, including how to access the raw data, preprocessed data, intermediate data, and generated data, etc.
        \item The authors should provide scripts to reproduce all experimental results for the new proposed method and baselines. If only a subset of experiments are reproducible, they should state which ones are omitted from the script and why.
        \item At submission time, to preserve anonymity, the authors should release anonymized versions (if applicable).
        \item Providing as much information as possible in supplemental material (appended to the paper) is recommended, but including URLs to data and code is permitted.
    \end{itemize}

\item {\bf Experimental setting/details}
    \item[] Question: Does the paper specify all the training and test details (e.g., data splits, hyperparameters, how they were chosen, type of optimizer) necessary to understand the results?
    \item[] Answer: \answerYes{} 
    \item[] Justification: The paper specifies the datasets, splits, evaluation metrics, training protocol, loss functions, and implementation details used in the experiments.
    \item[] Guidelines:
    \begin{itemize}
        \item The answer \answerNA{} means that the paper does not include experiments.
        \item The experimental setting should be presented in the core of the paper to a level of detail that is necessary to appreciate the results and make sense of them.
        \item The full details can be provided either with the code, in appendix, or as supplemental material.
    \end{itemize}

\item {\bf Experiment statistical significance}
    \item[] Question: Does the paper report error bars suitably and correctly defined or other appropriate information about the statistical significance of the experiments?
    \item[] Answer: \answerYes{} 
    \item[] Justification: We report the mean and standard deviation of the main REAL275 experiment over five random seeds in the appendix.
    \item[] Guidelines:
    \begin{itemize}
        \item The answer \answerNA{} means that the paper does not include experiments.
        \item The authors should answer \answerYes{} if the results are accompanied by error bars, confidence intervals, or statistical significance tests, at least for the experiments that support the main claims of the paper.
        \item The factors of variability that the error bars are capturing should be clearly stated (for example, train/test split, initialization, random drawing of some parameter, or overall run with given experimental conditions).
        \item The method for calculating the error bars should be explained (closed form formula, call to a library function, bootstrap, etc.)
        \item The assumptions made should be given (e.g., Normally distributed errors).
        \item It should be clear whether the error bar is the standard deviation or the standard error of the mean.
        \item It is OK to report 1-sigma error bars, but one should state it. The authors should preferably report a 2-sigma error bar than state that they have a 96\% CI, if the hypothesis of Normality of errors is not verified.
        \item For asymmetric distributions, the authors should be careful not to show in tables or figures symmetric error bars that would yield results that are out of range (e.g., negative error rates).
        \item If error bars are reported in tables or plots, the authors should explain in the text how they were calculated and reference the corresponding figures or tables in the text.
    \end{itemize}

\item {\bf Experiments compute resources}
    \item[] Question: For each experiment, does the paper provide sufficient information on the computer resources (type of compute workers, memory, time of execution) needed to reproduce the experiments?
    \item[] Answer: \answerYes{} 
    \item[] Justification: The paper reports the hardware used for training and evaluation, including GPU type and approximate training time.
    \item[] Guidelines:
    \begin{itemize}
        \item The answer \answerNA{} means that the paper does not include experiments.
        \item The paper should indicate the type of compute workers CPU or GPU, internal cluster, or cloud provider, including relevant memory and storage.
        \item The paper should provide the amount of compute required for each of the individual experimental runs as well as estimate the total compute. 
        \item The paper should disclose whether the full research project required more compute than the experiments reported in the paper (e.g., preliminary or failed experiments that didn't make it into the paper). 
    \end{itemize}
    
\item {\bf Code of ethics}
    \item[] Question: Does the research conducted in the paper conform, in every respect, with the NeurIPS Code of Ethics \url{https://neurips.cc/public/EthicsGuidelines}?
    \item[] Answer: \answerYes{} 
    \item[] Justification: The research uses standard public benchmarks and synthetic data generation, and does not involve human subjects, private data, or harmful data collection.
    \item[] Guidelines:
    \begin{itemize}
        \item The answer \answerNA{} means that the authors have not reviewed the NeurIPS Code of Ethics.
        \item If the authors answer \answerNo, they should explain the special circumstances that require a deviation from the Code of Ethics.
        \item The authors should make sure to preserve anonymity (e.g., if there is a special consideration due to laws or regulations in their jurisdiction).
    \end{itemize}

\item {\bf Broader impacts}
    \item[] Question: Does the paper discuss both potential positive societal impacts and negative societal impacts of the work performed?
    \item[] Answer: \answerYes{} 
    \item[] Justification: The paper discusses potential positive impacts in robotics and embodied perception, as well as possible limitations that may give rise to failures in real-world robotic systems.
    \item[] Guidelines:
    \begin{itemize}
        \item The answer \answerNA{} means that there is no societal impact of the work performed.
        \item If the authors answer \answerNA{} or \answerNo, they should explain why their work has no societal impact or why the paper does not address societal impact.
        \item Examples of negative societal impacts include potential malicious or unintended uses (e.g., disinformation, generating fake profiles, surveillance), fairness considerations (e.g., deployment of technologies that could make decisions that unfairly impact specific groups), privacy considerations, and security considerations.
        \item The conference expects that many papers will be foundational research and not tied to particular applications, let alone deployments. However, if there is a direct path to any negative applications, the authors should point it out. For example, it is legitimate to point out that an improvement in the quality of generative models could be used to generate Deepfakes for disinformation. On the other hand, it is not needed to point out that a generic algorithm for optimizing neural networks could enable people to train models that generate Deepfakes faster.
        \item The authors should consider possible harms that could arise when the technology is being used as intended and functioning correctly, harms that could arise when the technology is being used as intended but gives incorrect results, and harms following from (intentional or unintentional) misuse of the technology.
        \item If there are negative societal impacts, the authors could also discuss possible mitigation strategies (e.g., gated release of models, providing defenses in addition to attacks, mechanisms for monitoring misuse, mechanisms to monitor how a system learns from feedback over time, improving the efficiency and accessibility of ML).
    \end{itemize}
    
\item {\bf Safeguards}
    \item[] Question: Does the paper describe safeguards that have been put in place for responsible release of data or models that have a high risk for misuse (e.g., pre-trained language models, image generators, or scraped datasets)?
    \item[] Answer: \answerNA{}. 
    \item[] Justification: The paper does not release high-risk models, scraped datasets, or assets requiring special safeguards.
    \item[] Guidelines:
    \begin{itemize}
        \item The answer \answerNA{} means that the paper poses no such risks.
        \item Released models that have a high risk for misuse or dual-use should be released with necessary safeguards to allow for controlled use of the model, for example by requiring that users adhere to usage guidelines or restrictions to access the model or implementing safety filters. 
        \item Datasets that have been scraped from the Internet could pose safety risks. The authors should describe how they avoided releasing unsafe images.
        \item We recognize that providing effective safeguards is challenging, and many papers do not require this, but we encourage authors to take this into account and make a best faith effort.
    \end{itemize}

\item {\bf Licenses for existing assets}
    \item[] Question: Are the creators or original owners of assets (e.g., code, data, models), used in the paper, properly credited and are the license and terms of use explicitly mentioned and properly respected?
    \item[] Answer: \answerYes{} 
    \item[] Justification: The paper cites the existing datasets, tools, and codebases used in the work, including public benchmarks and rendering tools, and follows their intended research use.
    \item[] Guidelines:
    \begin{itemize}
        \item The answer \answerNA{} means that the paper does not use existing assets.
        \item The authors should cite the original paper that produced the code package or dataset.
        \item The authors should state which version of the asset is used and, if possible, include a URL.
        \item The name of the license (e.g., CC-BY 4.0) should be included for each asset.
        \item For scraped data from a particular source (e.g., website), the copyright and terms of service of that source should be provided.
        \item If assets are released, the license, copyright information, and terms of use in the package should be provided. For popular datasets, \url{paperswithcode.com/datasets} has curated licenses for some datasets. Their licensing guide can help determine the license of a dataset.
        \item For existing datasets that are re-packaged, both the original license and the license of the derived asset (if it has changed) should be provided.
        \item If this information is not available online, the authors are encouraged to reach out to the asset's creators.
    \end{itemize}

\item {\bf New assets}
    \item[] Question: Are new assets introduced in the paper well documented and is the documentation provided alongside the assets?
    \item[] Answer:  \answerYes{} 
    \item[] Justification: The paper introduces a schema-driven synthetic dataset/pipeline and documents its generation procedure, annotations, and intended research use. The release plan and documentation will be provided with the supplementary material or project page when applicable.
    \item[] Guidelines:
    \begin{itemize}
        \item The answer \answerNA{} means that the paper does not release new assets.
        \item Researchers should communicate the details of the dataset\slash code\slash model as part of their submissions via structured templates. This includes details about training, license, limitations, etc. 
        \item The paper should discuss whether and how consent was obtained from people whose asset is used.
        \item At submission time, remember to anonymize your assets (if applicable). You can either create an anonymized URL or include an anonymized zip file.
    \end{itemize}

\item {\bf Crowdsourcing and research with human subjects}
    \item[] Question: For crowdsourcing experiments and research with human subjects, does the paper include the full text of instructions given to participants and screenshots, if applicable, as well as details about compensation (if any)? 
    \item[] Answer: \answerNA{} 
    \item[] Justification: The paper does not involve crowdsourcing or research with human subjects.
    \item[] Guidelines:
    \begin{itemize}
        \item The answer \answerNA{} means that the paper does not involve crowdsourcing nor research with human subjects.
        \item Including this information in the supplemental material is fine, but if the main contribution of the paper involves human subjects, then as much detail as possible should be included in the main paper. 
        \item According to the NeurIPS Code of Ethics, workers involved in data collection, curation, or other labor should be paid at least the minimum wage in the country of the data collector. 
    \end{itemize}

\item {\bf Institutional review board (IRB) approvals or equivalent for research with human subjects}
    \item[] Question: Does the paper describe potential risks incurred by study participants, whether such risks were disclosed to the subjects, and whether Institutional Review Board (IRB) approvals (or an equivalent approval/review based on the requirements of your country or institution) were obtained?
    \item[] Answer: \answerNA{} 
    \item[] Justification: The paper does not involve human subjects research, so IRB approval is not applicable.
    \item[] Guidelines:
    \begin{itemize}
        \item The answer \answerNA{} means that the paper does not involve crowdsourcing nor research with human subjects.
        \item Depending on the country in which research is conducted, IRB approval (or equivalent) may be required for any human subjects research. If you obtained IRB approval, you should clearly state this in the paper. 
        \item We recognize that the procedures for this may vary significantly between institutions and locations, and we expect authors to adhere to the NeurIPS Code of Ethics and the guidelines for their institution. 
        \item For initial submissions, do not include any information that would break anonymity (if applicable), such as the institution conducting the review.
    \end{itemize}

\item {\bf Declaration of LLM usage}
    \item[] Question: Does the paper describe the usage of LLMs if it is an important, original, or non-standard component of the core methods in this research? Note that if the LLM is used only for writing, editing, or formatting purposes and does \emph{not} impact the core methodology, scientific rigor, or originality of the research, declaration is not required.
    \item[] Answer: \answerYes{} 
    \item[] Justification: LLM usage is declared in the appendix. 
    \item[] Guidelines:
    \begin{itemize}
        \item The answer \answerNA{} means that the core method development in this research does not involve LLMs as any important, original, or non-standard components.
        \item Please refer to our LLM policy in the NeurIPS handbook for what should or should not be described.
    \end{itemize}

\end{enumerate}

\end{document}